\documentclass[10pt,twocolumn,twoside]{IEEEtran}
\usepackage{setspace}
\usepackage{stfloats}
\usepackage{float}
\usepackage{cite}
\usepackage{psfig}
\usepackage{subfigure}
\usepackage[dvips]{graphicx}
\usepackage{amsmath}
\usepackage{amssymb}
\usepackage{multirow}
\newcommand{\bb}[1]{{\mathbf{#1}}}
\newcommand{\nn}{\nonumber}

\usepackage{mathtools}
\newlength{\arrow}
\newlength{\arrows}
\usepackage{color}

\begin{document}
\title{
Discrete Gyrator Transforms: Computational Algorithms and Applications}

\author{Soo-Chang Pei,~\IEEEmembership{Life Fellow,~IEEE}, Shih-Gu Huang, and Jian-Jiun Ding
\thanks{
Copyright (c) 2015 IEEE. Personal use of this material is permitted. However, permission to use this material for any other purposes must be obtained from the IEEE by sending a request to pubs-permissions@ieee.org.

This work was supported by the Ministry of Science and Technology, Taiwan, under Contract 103-2221-E-002-102.

S.~C. Pei and J.-J. Ding are with the Department of Electrical Engineering \& Graduate Institute of Communication Engineering, National Taiwan University, Taipei 10617, Taiwan (e-mail: pei@cc.ee.ntu.edu.tw, djj@cc.ee.ntu.edu.tw).

S.-G. Huang is with the Graduate Institute of Communication Engineering, National Taiwan University, Taipei 10617, Taiwan (e-mail: d98942023@ntu.edu.tw).}
}

\maketitle

\begin{abstract}
As an extension of the 2D fractional Fourier transform (FRFT) and a special case of the 2D linear canonical transform (LCT), the gyrator transform was introduced
to produce rotations in twisted space/spatial-frequency planes.
It is a useful tool in optics, signal processing and image processing.
In this paper, we
develop discrete gyrator transforms (DGTs) based on the 2D LCT.
Taking the advantage of the additivity property of the 2D LCT, we propose 
three kinds of DGTs,
each of which is a cascade of low-complexity operators.
These DGTs have different constraints, characteristics and properties, and are realized by different
computational algorithms.
Besides, we propose a kind of DGT based on the eigenfunctions of the gyrator transform.
This DGT
is an orthonormal transform, and thus its comprehensive properties, especially the additivity property,
make it more useful in many applications.
We also develop an efficient computational algorithm to significantly reduce the complexity of this DGT.
At the end, a brief review of some important applications of the DGTs is presented, including mode conversion, sampling and reconstruction, 
watermarking and image encryption.
\end{abstract}


\section{Introduction}\label{sec:intro}

Fractional Fourier transform (FRFT) \cite{namias1980fractional,ozaktas1993fourier,mendlovic1993fractional,ozaktas1993fractional,almeida1994fractional,pellat1994fresnel,alieva1994fractional}, as a generalization of the Fourier transform, is very useful in many applications such as optical system analysis, phase retrieval, filter design and pattern recognition.
The FRFT is a linear canonical integral transform that produces a rotation in the time/frequency plane $(x,\omega_x)$.
To extend the FRFT to two dimensions $(x,y)$, an easy and straightforward approach is performing two separate 1D FRFTs on two transverse directions, $x$ and $y$, respectively \cite{simon2000fractional}.
Accordingly, this 2D separable FRFT generates rotations in the space/spatial-frequency planes, $(x,\omega_x)$ and $(y,\omega_y)$.
In \cite{rodrigo2007gyrator}, another kind of 2D linear canonical integral transform, called gyrator transform, was proposed to produce rotations
in the twisted space/spatial-frequency planes, i.e. $(x,\omega_y)$ and $(y,\omega_x)$ planes.
Given a 2D signal $g(x,y)$, the gyrator transform with rotation
angle $\alpha$ is
\begin{align}\label{eq:intro04}
&G(u,v) = {\rm GT}_\alpha\left\{ {g(x,y)} \right\}= \frac{{\left| {\csc \alpha } \right|}}{{2\pi }}
\nn\\
&\ \cdot\int\limits_{ - \infty }^\infty  {\!\!\int\limits_{ - \infty }^\infty  \!\!{\exp \left[ {\frac{{j\left( {uv + xy} \right)}}{{\tan \alpha }} - \frac{{j(uy + vx)}}{{\sin \alpha }}} \right]} } g(x,y)dxdy.
\end{align}
It is obvious that the above definition is singular at $\alpha= k\pi$.
When
$\alpha=2k\pi$, the gyrator transform is defined as $G(u,v)=g(u,v)$; and when
$\alpha=(2k+1)\pi$, $G(u,v)=g(-u,-v)$.
If $G_1(\omega_x,\omega_y)$ is defined as the 2D Fourier transform of $g(x,y)$, the gyrator transform with $\alpha=\pi/2$ reduces to the reflection of $G_1(\omega_x,\omega_y)$, i.e. $G(u,v)=G_1(v,u)$.
The gyrator transform cannot be separated into two 1D transforms, and thus it is sometimes classified as a kind of 2D nonseparable FRFT.

In \cite{rodrigo2007gyrator}, the optical
implementation of the gyrator transform has been discussed. And several properties of the gyrator transform have been derived in \cite{rodrigo2007gyrator,pei2009properties}.
The focus of this paper is on the digital implementations of the gyrator transform, called discrete gyrator transforms (DGTs) for short.
Suppose the sampling intervals in space domain and spatial-frequency domain are $(\Delta_x,\Delta_y)$ and $(\Delta_u,\Delta_v)$, respectively:
\begin{align}\label{eq:intro06}
g[m,n] \buildrel \Delta \over = g\left(m{\Delta _x},n{\Delta _y}\right),\quad G[p,q] \buildrel \Delta \over = G\left(p{\Delta _u},q{\Delta _v}\right).
\end{align}
The simplest way to derive the DGT is sampling the continuous gyrator transform and computing it directly by summation:
\begin{align}\label{eq:intro08}
G[p,q] &= {\rm DGT}_\alpha \left\{ {g[m,n]} \right\}\nn\\
& = \frac{{\left| {\csc \alpha } \right|}}{{2\pi }}\sum\limits_m^{} {\sum\limits_n^{} {\exp \left[ {\frac{{j\left( {pq{\Delta _u}{\Delta _v} + mn{\Delta _x}{\Delta _y}} \right)}}{{\tan \alpha }}} \right.} } \nn\\
&\quad\ \ \left.{ - \frac{{j\left( {pn{\Delta _u}{\Delta _y} + qm{\Delta _v}{\Delta _x}} \right)}}{{\sin \alpha }}} \right]g[m,n]{\Delta _x}{\Delta _y}.
\end{align}
The advantage of this discrete transform is that there are no constraints on $(\Delta_x,\Delta_y)$ and $(\Delta_u,\Delta_v)$, but it has very high computational complexity and is thus time-consuming.

In \cite{pei2009properties,liu2011fast}, some low-complexity DGTs implemented by discrete Fourier transform (DFT) or
convolution were proposed.
These DGTs are derived directly from (\ref{eq:intro04}) and (\ref{eq:intro08}).
In this paper, we develop DGTs from the point of view of 2D linear canonical transform (LCT).
The gyrator transform is a special case of the 2D LCT.
Using the additivity property of the 2D LCT, the gyrator transform can be factorized into a sequence of low-complexity transforms.
With a different decomposition method, a different DGT can be developed.
In this paper, three kinds of DGTs are proposed based on the 2D LCT.
The first one is realized by 2D linear convolution, the second uses the 2D DFT, and the last one is implemented by 2D circular convolution.
Since different computational algorithms are utilized, they have different constraints on the sampling intervals, different characteristics and properties, and different computational complexity.
The DGTs in \cite{pei2009properties,liu2011fast} are the special cases of the proposed DGTs.
The first two proposed DGTs are singular at $\alpha=k\pi$, while the third one is singular at $\alpha=(2k+1)\pi$.
When $\alpha$ is close to $k\pi$ or $(2k+1)\pi$, these DGTs suffer from low-accuracy and overlapping (aliasing) problems.
Accordingly, a method is proposed to help the DGTs avoid these problems.

The DGTs mentioned above have unitary and reversibility properties.
However, they don't satisfy the additivity property,
which is useful in many signal/image processing applications.
Accordingly, we develop the 4th kind of DGT, which is based on the eigenfunctions of the gyrator transform.
It has been shown in \cite{pei2009properties} that rotated Hermite Gaussian functions (RHGFs) are the eigenfunctions of the continuous gyrator transform.
For the discrete case, we generate discrete orthonormal RHGFs from 1D discrete Hermite Gaussian functions (HGFs) given by \cite{pei2008generalized}.
The DGT based on the discrete HGFs is an orthonormal transform, and therefore it satisfies many properties
including unitary, reversibility and additivity.
To reduce the complexity of this DGT, we also develop an efficient computational algorithm.
In the end of this paper, to emphasize the importance of the proposed DGTs, some applications
are introduced, including mode conversion, sampling and reconstruction, 
watermarking and image encryption.

\section{
Development of Discrete Gyrator Transforms Based on 2D Linear Canonical Transform}\label{sec:DGTlct}
In this section, we develop DGTs from the 
2D LCT.
The 2D LCT \cite{folland1989harmonic,sahin1995optical,pei2001two} with parameter matrix $\bb M$, denoted by ${\rm{LCT}}_{\bf{M}}$, is an affine transform with ten degrees of freedom,
\begin{align}\label{eq:DGTlct04}
G({\bf{u}}) &= {\rm{LCT}}_{\bf{M}}\left\{ {g({\bf{x}})} \right\}\nn\\
& = \frac{1}{{2\pi \sqrt { - \det ({\bf{B}})} }}\!\int\! \exp \left[ \frac{j}{2}\left( {{\bf{u}}^T}{\bf{D}}{{\bf{B}}^{ - 1}}{\bf{u}} - 2{{\bf{x}}^T}{{\bf{B}}^{ - 1}}{\bf{u}} \right.\right.\nn\\
&\qquad\qquad\qquad\qquad\quad\ \  \left.\left.\frac{}{}+ {{\bf{x}}^T}{{\bf{B}}^{ - 1}}{\bf{Ax}} \right) \right]g({\bf{x}})d{\bf{x}},
\end{align}
where $\bb x=[x,y]^T$ and $\bb u=[u,v]^T$. The
$4\times4$ parameter matrix $\bb M$ is defined as $\bb M=[\bb A, \bb B; \bb C, \bb D]$, where $\bb A$, $\bb B$, $\bb C$ and $\bb D$ are $2\times2$ matrices satisfying
\begin{align}\label{eq:DGTlct06}
{{\bf{A}}^T}{\bf{C}} = {{\bf{C}}^T}{\bf{A}},\ {{\bf{B}}^T}{\bf{D}} = {{\bf{D}}^T}{\bf{B}},\ {{\bf{A}}^T}{\bf{D}} - {{\bf{C}}^T}{\bf{B}} = {\bf{I}}.
\end{align}
Suppose the spatial-frequency coordinates with respect to $(x,y)$ and $(u,v)$ are $(\omega_x,\omega_y)$ and $(\omega_u,\omega_v)$, respectively.
The gyrator transform is a special case of the 2D LCT that performs rotations in the $(x,\omega_y)$ and $(y,\omega_x)$ planes.
That is,
\begin{align}\label{eq:DGTlct08}
\begin{bmatrix}
u\\
v\\
{{\omega _u}}\\
{{\omega _v}}
\end{bmatrix}
= \underbrace{\begin{bmatrix}
{\cos \alpha }&0&0&{\sin \alpha }\\
0&{\cos \alpha }&{\sin \alpha }&0\\
0&{ - \sin \alpha }&{\cos \alpha }&0\\
{ - \sin \alpha }&0&0&{\cos \alpha }
\end{bmatrix}}_{\bb M_\alpha}
\begin{bmatrix}
x\\
y\\
{{\omega _x}}\\
{{\omega _y}}
\end{bmatrix}.
\end{align}
Denote the above $4\times4$ matrix as $\bb M_\alpha$.
If we let $\bb M=\bb M_\alpha$ in the 2D LCT in (\ref{eq:DGTlct04}),
the 2D LCT becomes the gyrator transform.

The 2D LCT satisfies the additivity property, i.e.
\begin{align}\label{eq:DGTlct10}
{\rm{LC}}{{\rm{T}}_{{{\widetilde {\bf{M}}}_1}}}{\rm{LC}}{{\rm{T}}_{{{\widetilde {\bf{M}}}_2}}} = {\rm{LC}}{{\rm{T}}_{{{\widetilde {\bf{M}}}_1} \times {{\widetilde {\bf{M}}}_2}}}.
\end{align}
If we decompose the parameter matrix $\bb M_\alpha$ into $k$ matrices,
\begin{align}\label{eq:DGTlct12}
{{\bf{M}}_\alpha } = {\widetilde {\bf{M}}_k} \times {\widetilde {\bf{M}}_{k - 1}} \times  \cdots  \times {\widetilde {\bf{M}}_1},
\end{align}
the gyrator transform, denoted by ${\rm{G}}{{\rm{T}}_\alpha }$, can be realized by a sequence of $k$ 2D LCTs, i.e.
\begin{align}\label{eq:DGTlct14}
{\rm{G}}{{\rm{T}}_\alpha } = {\rm{LC}}{{\rm{T}}_{{{\bf{M}}_\alpha }}}{\rm{ = LC}}{{\rm{T}}_{{{\widetilde {\bf{M}}}_k}}}{\rm{LC}}{{\rm{T}}_{{{\widetilde {\bf{M}}}_{k - 1}}}} \cdots {\rm{LC}}{{\rm{T}}_{{{\widetilde {\bf{M}}}_1}}}.
\end{align}
In order to achieve low complexity for digital implementation,
we require each of the $k$ transforms
to be a simple 2D  operator such as a reflection, multiplication, convolution or Fourier transform.
If so, a DGT can be designed  as a sequence of low-complexity discrete transforms.
In the following,
three kinds of DGTs are developed based
on (\ref{eq:DGTlct12}) and (\ref{eq:DGTlct14}),
and we will show that the DGTs in \cite{pei2009properties,liu2011fast} are the special cases of the proposed DGTs.
Some important properties, constraints and comparisons of these DGTs will also be discussed.

\subsection{DGT Based On Linear Chirp Convolution (DGT-LCC)}\label{subsec:DGT1cc1}
Suppose the parameter matrix corresponding to the gyrator transform in (\ref{eq:DGTlct08}) is decomposed as
\begin{align}\label{eq:DGT1cc104}
\mathbf{M}_{\alpha} =&
\underbrace{\begin{bmatrix}
1&0&0&0\\
0&1&0&0\\
{ - \csc \alpha }&{\cot \alpha }&1&0\\
{\cot \alpha }&{ - \csc \alpha }&0&1
\end{bmatrix}}_{(\ref{eq:DGT1cc106_1})}
\underbrace{\begin{bmatrix}
0&1&0&0\\
1&0&0&0\\
0&0&0&1\\
0&0&1&0
\end{bmatrix}}_{(\ref{eq:DGT1cc106_2})}\nn\\
&\quad\times
\underbrace{\begin{bmatrix}
1&0&{\sin \alpha }&0\\
0&1&0&{\sin \alpha }\\
0&0&1&0\\
0&0&0&1
\end{bmatrix}}_{(\ref{eq:DGT1cc106_3})}
\underbrace{\begin{bmatrix}
1&0&0&0\\
0&1&0&0\\
{ - \csc \alpha }&{\cot \alpha }&1&0\\
{\cot \alpha }&{ - \csc \alpha }&0&1
\end{bmatrix}}_{(\ref{eq:DGT1cc106_4})},
\end{align}
where the index under each matrix shows the equation number of the corresponding 2D operator.
In the 2D LCT, these four matrices in turn
(from right to left) correspond to 2D chirp multiplication,  chirp convolution, reflection, and again the same chirp multiplication.
Therefore, the gyrator transform can be expressed as a sequence of the following four 2D operators:
\begin{align}
&\!\!\!{g_1}\!(x,y\!)\! = \!\exp \!\!\left[ \!{ - \frac{j}{2}({x^2} + {y^2})\csc \alpha} \right]\!\!\exp (jxy\cot \alpha \!)g(x,y),\label{eq:DGT1cc106_4}\\
&\!\!\!{G_1}(v,u) =  \frac{{|\csc \alpha |}}{{2\pi }}\!\!\!\!\int\limits_{ - \infty }^\infty \!\!\int\limits_{ - \infty }^\infty  \!\! {\exp \left[ {\frac{{j\csc \alpha }}{2}\left( {{{(v - x)}^2}} \right.} \right.}   \nn\\
&\qquad\qquad\qquad\qquad \qquad \left. {\frac{{}}{{}}\left. +{{{(u - y)}^2}} \right)} \right]{g_1}(x,y)dxdy,\label{eq:DGT1cc106_3}\\
&\!\!\!{G_2}(u,v) = {G_1}(v,u), \label{eq:DGT1cc106_2}\\
&\!\!\!G\!(u,v\!) \!\!= \!\exp \!\!\left[ \!{ - \frac{j}{2}({u^2} + {v^2})\csc \alpha }\! \right]\!\!\exp (juv\cot \alpha \!){G_2}\!(u,v).\label{eq:DGT1cc106_1}
\end{align}
In the discrete case,
assume the sampling intervals of $x,y,u,v$ are $\Delta_x,\Delta_y, \Delta_u,\Delta_v$, respectively, and the discrete input is $g[m,n]=g(m\Delta_x,n\Delta_y)$.
Let
\begin{align}\label{eq:DGT1cc109_1}
(v - x)^2&=q^2\Delta_v^2 + m^2\Delta_x^2-2qm\Delta_v\Delta_x\nn\\
&=(q-m)^2\Delta_v\Delta_x+q^2\left(\Delta_v^2-\Delta_v\Delta_x\right)\nn\\
&\qquad\qquad\qquad\qquad+ m^2\left(\Delta_x^2-\Delta_v\Delta_x\right),\\
(u - y)^2&=(p-n)^2\Delta_u\Delta_y+p^2\left(\Delta_u^2-\Delta_u\Delta_y\right)\nn\\
&\qquad\qquad\qquad\qquad+ n^2\left(\Delta_y^2-\Delta_u\Delta_y\right).\label{eq:DGT1cc109_2}
\end{align}
It can be found that there are no constraints on $\Delta_x,\Delta_y,\Delta_u,\Delta_v$ in the above digital implementations.
From (\ref{eq:DGT1cc109_1}) and (\ref{eq:DGT1cc109_2}),
the DGT based on (\ref{eq:DGT1cc106_4})-(\ref{eq:DGT1cc106_1}) is given by
\begin{align}
&{g_1}[m,n] = \exp \left[ { - \frac{j}{2}({m^2}\Delta _v\Delta _x + {n^2}\Delta _u\Delta _y)\csc \alpha } \right]\nn\\
&\qquad\qquad\qquad\qquad\quad\cdot\exp (jmn{\Delta _x}{\Delta _y}\cot \alpha )g[m,n],\label{eq:DGT1cc110_1}\\
&{G_1}[q,p] = \frac{{|\csc \alpha |{\Delta _x}{\Delta _y}}}{{2\pi }}\!\sum\limits_m  \!\sum\limits_n \! {\exp \left[ {\frac{{j }}{2}\left( {(q - m)^2\Delta _v\Delta _x} \right.} \right.}  \nn\\
&\qquad\qquad\qquad\quad\ \left. {\frac{{}}{{}}\left. { + (p - n)^2\Delta _u\Delta _y} \right)\csc \alpha} \right]{g_1}[m,n],\label{eq:DGT1cc110_2}\\
&G[p,q] = \exp \left[ { - \frac{j}{2}({p^2}\Delta _u\Delta _y + {q^2}\Delta _v\Delta _x)\csc \alpha } \right]\nn\\
&\qquad\qquad\qquad\qquad\quad\ \cdot\exp(jpq{\Delta _u}{\Delta _v}\cot \alpha ){G_1}[q,p].\label{eq:DGT1cc110_3}
\end{align}
The first step (\ref{eq:DGT1cc110_1}) corresponds to (\ref{eq:DGT1cc106_4}), the second step (\ref{eq:DGT1cc110_2}) to (\ref{eq:DGT1cc106_3}), and the third step (\ref{eq:DGT1cc110_3}) to the combination of (\ref{eq:DGT1cc106_2}) and (\ref{eq:DGT1cc106_1}).
The key feature of this DGT is the use of linear chirp convolution (LCC), and thus it is called DGT based on LCC (DGT-LCC).
The Method 2 in \cite{pei2009properties} is the special case of the DGT-LCC that $\Delta_x=\Delta_y$ and $\Delta_u=\Delta_v$ are used.

The linear chirp convolution in (\ref{eq:DGT1cc110_2}) can be efficiently calculated by
2D FFT algorithm, i.e. three 2D fast Fourier transforms (FFTs) and one
pointwise product.
The chirp function in (\ref{eq:DGT1cc110_2}) is truncated when calculating its 2D FFT.
For example, if the size of input $g_1$ is $N_1\times N_2$ and we want to obtain $N_1\times N_2$ output data without truncation error, the chirp function should be of size $(2N_1-1)\times(2N_2-1)$.
And it follows that the whole output of the linear convolution is of size $(3N_1-2)\times(3N_2-2)$.
Although only the central $N_1\times N_2$ output data are without truncation error, the rest must be retained for lossless recovery.

\begin{figure}[t]
\centering
\includegraphics[width=.95\columnwidth,clip=true]{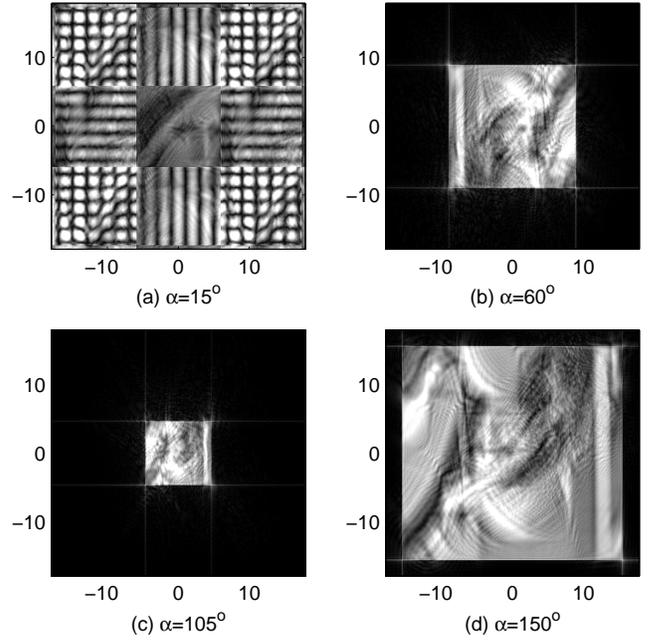}
\caption{
Magnitudes of the DGT-LCCs of a $512\times512$ two-times upsampled Lena image: (a) $\alpha=15^o$, (b) $\alpha=60^o$, (c) $\alpha=105^o$, and (d) $\alpha=150^o$,  where ${\Delta _x}={\Delta _y}={\Delta _u}={\Delta _v}= 0.07$.
(The output size is larger then the input size because of linear convolution.
Only the central $512\times512$ output data are shown here.)
}
\label{fig:DGT1cc1}
\end{figure}

It is obvious that the DGT-LCC is not suitable for
$\alpha\to k\pi$ because $\cot\alpha, \csc\alpha\to\pm\infty$ in (\ref{eq:DGT1cc110_1})-(\ref{eq:DGT1cc110_3}).
The accuracy of the DGT-LCC will decrease because the absolute values of $\cot\alpha$ and $\csc\alpha$ are too large to be accurately described in practical implementation.
Besides, the high chirp rate in the chirp multiplication in (\ref{eq:DGT1cc110_1}) yields a substantial shearing in spatial-frequency domain and subsequent larger bandwidth.
If the sampling intervals $\Delta_x$ and $\Delta_y$ are not small enough, overlapping (aliasing) effect will be produced.
For example, consider the input is a $256\times256$ Lena image with $\Delta_x=\Delta_y=0.14$.
The DGT-LCCs with $\alpha$ being $15^o$, $60^o$, $105^o$ and $150^o$ are examined.
Obvious overlapping (aliasing) effect occurs when $\alpha=15^o$ and $\alpha=150^o$.
To reduce the overlapping (aliasing) effect, the input image is two-times upsampled to $512\times512$ so that $\Delta_x$ and $\Delta_y$ decrease to $0.07$.
The DGT-LCCs with $\Delta_u=\Delta_v=0.07$ are displayed in Fig.~\ref{fig:DGT1cc1}.
It is shown that the $150^o$ case is out of overlapping (aliasing) problem, but the $\alpha=15^o$ case isn't because it is much closer to $k\pi$ and requires much smaller $\Delta_x$ and $\Delta_y$.

\subsection{DGT Based On Discrete Fourier Transform (DGT-DFT)}\label{subsec:DGT1dft}
The parameter matrix
$\mathbf{M}_{\alpha}$ in (\ref{eq:DGTlct08}) can also be factorized into:
\begin{align}\label{eq:DGT1dft04}
\mathbf{M}_{\alpha} =&
\underbrace{\begin{bmatrix}
1&0&0&0\\
0&1&0&0\\
0&{\cot \alpha }&1&0\\
{\cot \alpha }&0&0&1
 \end{bmatrix}}_{(\ref{eq:DGT1dft06_1})}
\underbrace{\begin{bmatrix}
0&1&0&0\\
1&0&0&0\\
0&0&0&1\\
0&0&1&0
\end{bmatrix}}_{(\ref{eq:DGT1dft06_2})}
\underbrace{\begin{bmatrix}
{\sin \alpha }&0&0&0\\
0&{\sin \alpha }&0&0\\
0&0&{\csc \alpha }&0\\
0&0&0&{\csc \alpha }
 \end{bmatrix}}_{(\ref{eq:DGT1dft06_3})}\nn\\
&\qquad\qquad\quad\times
\underbrace{\begin{bmatrix}
0&0&1&0\\
0&0&0&1\\
{ - 1}&0&0&0\\
0&{ - 1}&0&0
 \end{bmatrix}}_{(\ref{eq:DGT1dft06_4})}
\underbrace{\begin{bmatrix}
1&0&0&0\\
0&1&0&0\\
0&{\cot \alpha }&1&0\\
{\cot \alpha }&0&0&1
 \end{bmatrix}}_{(\ref{eq:DGT1dft06_5})}.
\end{align}
The 2D LCTs with these five matrices
(from right to left) are respectively equivalent to 2D chirp multiplication, Fourier transform, scaling, reflection, and again the same chirp multiplication.
It implies that the gyrator transform can be expressed as the cascade of the five 2D operators below:
\begin{align}
{g_1}(x,y) &= \exp (jxy\cot \alpha )g(x,y),\label{eq:DGT1dft06_5}\\
&\hspace{-1.2cm}{G_1}(v\csc \alpha ,u\csc \alpha )=  \frac{1}{2\pi}\!\int\limits_{ - \infty }^\infty  \!\int\limits_{ - \infty }^\infty  \!\!  \exp \left[ - j(v\csc \alpha)x \right.\nn\\
&\qquad\qquad\qquad\quad \left.- j(u\csc \alpha)y\right]{g_1}(x,y)  dxdy,\label{eq:DGT1dft06_4}\\
{G_2}(v,u) &= |\csc \alpha |{G_1}(v\csc \alpha ,u\csc \alpha ),\label{eq:DGT1dft06_3}\\
{G_3}(u,v) &= {G_2}(v,u),\label{eq:DGT1dft06_2}\\
G(u,v) &= \exp (juv\cot \alpha ){G_3}(u,v).\label{eq:DGT1dft06_1}
\end{align}
Consider a discrete input $g[m,n]=g(m\Delta_x,n\Delta_y)$ of size $N_1\times N_2$.
In order to realize (\ref{eq:DGT1dft06_4})
by  DFT or inverse DFT (IDFT), the
requirements are
\begin{align}\label{eq:DGT1dft08}
{\Delta _x}{\Delta _v} = \frac{{2\pi | {\sin \alpha }|}}{N_1}, \quad
{\Delta _y}{\Delta _u} = \frac{{2\pi | {\sin \alpha }|}}{N_2}.
\end{align}
Then the
discrete output $G[p,q]=G(p\Delta_u,q\Delta_v)$ can be obtained from the DGT defined as the following three steps:
\begin{align}
&{g_1}[m,n] = \exp (jmn{\Delta _x}{\Delta _y}\cot \alpha )g[m,n],\label{eq:DGT1dft10_1}\\
&{G_1}[q,p] \!=\! \frac{{\Delta _x}{\Delta _y}}{2\pi} \!\sum\limits_m  \!\sum\limits_n \! {\exp \left( \mp  j\frac{{2\pi qm }}{N_1} \mp  j\frac{{2\pi pn}}{N_2}\right)\!{g_1}[m,n]} ,\label{eq:DGT1dft10_2}\\
&G[p,q] = |\csc\alpha|\exp (jpq{\Delta _u}{\Delta _v}\cot \alpha ){G_1}[q,p]\label{eq:DGT1dft10_3}.
\end{align}
The first step (\ref{eq:DGT1dft10_1}) corresponds to (\ref{eq:DGT1dft06_5}), the second step (\ref{eq:DGT1dft10_2}) to (\ref{eq:DGT1dft06_4}), and the third step (\ref{eq:DGT1dft10_3}) to the combination of (\ref{eq:DGT1dft06_3})-(\ref{eq:DGT1dft06_1}).
For the two minus-plus signs $\mp$ in (\ref{eq:DGT1dft10_2}), minus is used when $\sin\alpha>0$ while plus is used when $\sin\alpha<0$.
Since this DGT is carried out by the DFT/IDFT, it is called DGT
based on DFT (DGT-DFT).
When $N_1=N_2$, $\Delta_x=\Delta_y$ and $\sin\alpha>0$, the DGT-DFT is equivalent to Method 1 in \cite{pei2009properties}.
For the fast algorithm of the DGT-DFT, one 2D FFT is utilized for the 2D DFT/IDFT in (\ref{eq:DGT1dft10_2}) and dominates the complexity.

\begin{figure}[t]
\centering
\includegraphics[width=.95\columnwidth,clip=true]{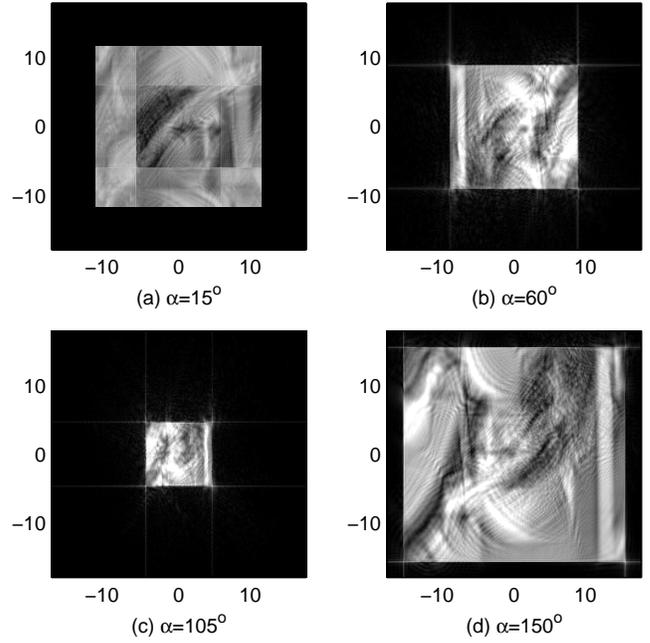}
\caption{
Magnitudes of the DGT-DFTs of a $512\times512$ two-times upsampled Lena image: (a) $\alpha=15^o$, (b) $\alpha=60^o$, (c) $\alpha=105^o$, and (d) $\alpha=150^o$, where ${\Delta _x}={\Delta _y}= 0.07$ and ${\Delta _u}={\Delta _v}= 2\pi|\sin\alpha|/512/0.07$.
(For different $\alpha$'s, the ranges of $u$ and $v$ are different because different $\Delta_u$ and $\Delta_v$ are used.) 
}
\label{fig:DGT1dft}
\end{figure}

Like the DGT-LCC, the DGT-DFT also suffers from low-accuracy and overlapping (aliasing) problems when
$\alpha\to k\pi$.
Again, using the $256\times256$ Lena image with $\Delta_x=\Delta_y=0.14$ as the input, the
cases of $\alpha$ being $15^o$, $60^o$, $105^o$ and $150^o$ are analyzed.
The DGT-DFTs with $\alpha=15^o$ and $\alpha=150^o$ have severe overlapping (aliasing) problem.
If the sampling interval $\Delta_x$ and $\Delta_y$ is reduced to 0.07 by two-times upsampling, the resulting DGT-DFTs in Fig.~\ref{fig:DGT1dft} show that the overlapping (aliasing) effect in the $150^o$ case is eliminated.
But $\Delta_x$ and $\Delta_y$ are still not small enough for the $15^o$ case, which is closer to $k\pi$ then the $150^o$ case.
Note that the output sampling intervals depend on $\alpha$, according to the constraints given in (\ref{eq:DGT1dft08}).


\subsection{DGT Based On Circular Chirp Convolution (DGT-CCC)}\label{subsec:DGT1cc2}
If the following constraints are used:
\begin{align}\label{eq:DGT1cc202}
{\Delta _u}={\Delta _x}, \quad
{\Delta _v}={\Delta _y},
\end{align}
the DGT-LCC can reduce to the DGT based on the following more concise decomposition:
\begin{align}\label{eq:DGT1cc204}
\mathbf{M}_{\alpha} &=
\underbrace{\begin{bmatrix}
1&0&0&0\\
0&1&0&0\\
0&{ - \tan \frac{\alpha }{2}}&1&0\\
{ - \tan \frac{\alpha }{2}}&0&0&1
\end{bmatrix}}_{(\ref{eq:DGT1cc216_5})}
\begin{bmatrix}
1&0&0&{\sin \alpha }\\
0&1&{\sin \alpha }&0\\
0&0&1&0\\
0&0&0&1
\end{bmatrix}\nn\\
&\qquad\qquad\qquad\quad\ \times
\underbrace{\begin{bmatrix}
1&0&0&0\\
0&1&0&0\\
0&{ - \tan \frac{\alpha }{2}}&1&0\\
{ - \tan \frac{\alpha }{2}}&0&0&1
\end{bmatrix}}_{(\ref{eq:DGT1cc216_1})}.
\end{align}
In the 2D LCT, these three matrices correspond to 2D chirp multiplication, chirp convolution and the same chirp multiplication again, respectively.
However, like the DGT-LCC, this DGT also has the disadvantage that the output size is larger than the input size due to the linear convolution.
Fortunately, if (\ref{eq:DGT1cc202}) is satisfied, this disadvantage can be avoided by replacing the linear convolution by circular convolution.

The second matrix in (\ref{eq:DGT1cc204}) can be further decomposed into
\begin{align}\label{eq:DGT1cc212}
\underbrace{\begin{bmatrix}
0&0&{-1}&0\\
0&0&0&{-1}\\
  1&0&0&0\\
0&  1&0&0
 \end{bmatrix}}_{(\ref{eq:DGT1cc216_4})}
\underbrace{\begin{bmatrix}
1&0&0&0\\
0&1&0&0\\
0&{ - \sin \alpha }&1&0\\
{ - \sin \alpha }&0&0&1
\end{bmatrix}}_{(\ref{eq:DGT1cc216_3})}
\underbrace{\begin{bmatrix}
0&0&1&0\\
0&0&0&1\\
{ - 1}&0&0&0\\
0&{ - 1}&0&0
 \end{bmatrix}}_{(\ref{eq:DGT1cc216_2})}.
\end{align}
The above three matrices from right to left represent 2D Fourier transform, chirp multiplication, and inverse Fourier transform, respectively.
The decompositions (\ref{eq:DGT1cc204}) and (\ref{eq:DGT1cc212}) show that the gyrator transform can be expressed as the cascade of the following five 2D operators:
\begin{align}
&{g_1}(x,y) = \exp \left(-jxy\tan \frac{\alpha}{2} \right)g(x,y),\label{eq:DGT1cc216_1}\\
&G_1(x',y') \!= \frac{{ 1}}{{2\pi }}\!\!\!\int\limits_{ - \infty }^\infty \!  \int\limits_{ - \infty }^\infty  \!\!\!\exp ( - jx'x - jy'y)g_1(x,y)dxdy,\label{eq:DGT1cc216_2}\\
&G_2(x',y') = \exp (-jx'y'\sin\alpha )G_1(x',y'),\label{eq:DGT1cc216_3}\\
&g_2(u,v) \!= \frac{{ 1}}{{2\pi }}\!\!\int\limits_{ - \infty }^\infty \!  \int\limits_{ - \infty }^\infty  \!\!\! \exp (jux' +jvy')G_2(x',y')dx'dy',\label{eq:DGT1cc216_4}\\
&G(u,v) =\exp \left(-juv\tan \frac{\alpha}{2} \right){g_2}(u,v).\label{eq:DGT1cc216_5}
\end{align}
Assume the discrete input is of size $N_1\times N_2$.
If we 
realize (\ref{eq:DGT1cc216_2}) by 2D DFT and (\ref{eq:DGT1cc216_4}) by 2D IDFT, the sampling intervals for $x'$ and $y'$, denoted by $\Delta_{x'}$ and $\Delta_{y'}$, are set to satisfy $\Delta_{x'}\Delta_x=\Delta_u\Delta_{x'}=2\pi/N_1$ and  $\Delta_{y'}\Delta_y=\Delta_v\Delta_{y'}=2\pi/N_2$.
This also explains why the constraints ${\Delta _u}={\Delta _x}$ and ${\Delta _v}={\Delta _y}$ in (\ref{eq:DGT1cc202}) are necessary.
The DGT based on (\ref{eq:DGT1cc216_1})-(\ref{eq:DGT1cc216_5}) is given by
\begin{align}
&{g_1}[m,n] = \exp \left( { - jmn{\Delta _x}{\Delta _y}\tan \frac{\alpha }{2}} \right)g[m,n],\label{eq:DGT1cc220_1}\\
&G_1[m',n'] = \frac{{{\Delta _x}{\Delta _y}}}{{2\pi }}\sum\limits_m  \sum\limits_n  \exp \left( { - j\frac{{2\pi m'm}}{{{N_1}}} - j\frac{{2\pi n'n}}{{{N_2}}}} \right)\nn\\
&\qquad\qquad\qquad\qquad\qquad\qquad\qquad\qquad\qquad\cdot g_1[m,n],\label{eq:DGT1cc220_2}\\
&G_2[m',n']=\exp( - jm'n'{\Delta _{x'}}{\Delta _{y'}}\sin \alpha )G_1[m',n'],\label{eq:DGT1cc220_3}\\
&{g_2}[p,q] = \frac{{{\Delta _{x'}}{\Delta _{y'}}}}{{2\pi }}\sum\limits_{m'} {\sum\limits_{n'} {\exp } } \left( { + j\frac{{2\pi pm'}}{{{N_1}}} + j\frac{{2\pi qn'}}{{{N_2}}}} \right)\nn\\
&\qquad\qquad\qquad\qquad\qquad\qquad\qquad\qquad\quad\cdot G_2[m',n'],\label{eq:DGT1cc220_4}\\
&G[p,q] = \exp \left( { - jpq{\Delta _u}{\Delta _v}\tan \frac{\alpha }{2}} \right){g_2}[p,q].\label{eq:DGT1cc220_5}
\end{align}
The main feature of this DGT is the circular convolution with a chirp function, i.e. (\ref{eq:DGT1cc220_2})-(\ref{eq:DGT1cc220_4}).
Thus, it is called DGT based on circular chirp convolution (DGT-CCC).
The DGT proposed in \cite{liu2011fast} is a special case of the DGT-CCC where ${\Delta _u}={\Delta _x}=\sqrt{2\pi/N_1}$ and ${\Delta _v}={\Delta _y}=\sqrt{2\pi/N_2}$.

\begin{figure}[t]
\centering
\includegraphics[width=.95\columnwidth,clip=true]{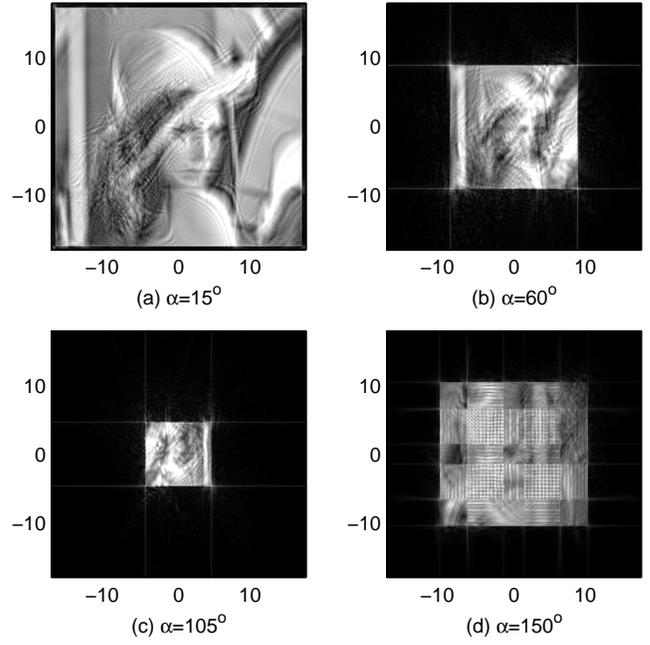}
\caption{
Magnitudes of the DGT-CCCs of a $512\times512$ two-times upsampled Lena image: (a) $\alpha=15^o$, (b) $\alpha=60^o$, (c) $\alpha=105^o$, and (d) $\alpha=150^o$, where ${\Delta _x}={\Delta _y}={\Delta _u}={\Delta _v}= 0.07$
}
\label{fig:DGT1cc2}
\end{figure}

The dominant complexity of the DGT-CCC is on the two 2D FFTs used for the 2D DFT in (\ref{eq:DGT1cc220_2}) and 2D IDFT in (\ref{eq:DGT1cc220_4}).
Unlike the DGT-LCC and DGT-DFT, the DGT-CCC is singular only at
$\alpha=(2k+1)\pi$ because $\tan\frac{\alpha}{2}\to\pm\infty$.
When $\alpha$ is closer to
$(2k+1)\pi$, the DGT-CCC suffers from more severe low-accuracy and overlapping (aliasing) problems.
For example, repeat the simulation in Fig.~\ref{fig:DGT1cc1} except that the
DGT-CCC is employed.
When $\Delta_x=\Delta_y=0.14$, the DGT-CCC performs well for $\alpha=15^o$ and $60^o$, but produces overlapping (aliasing) effect for $\alpha=105^o$ and $150^o$.
If the input is two-times upsampled to $512\times512$ ($\Delta_x$ and $\Delta_y$ becomes 0.07), the DGT-CCC with $\alpha=105^o$ doesn't have overlapping (aliasing) problem anymore, as shown in Fig.~\ref{fig:DGT1cc2}.
However, $\Delta_x=\Delta_y=0.07$ is still not small enough for $\alpha=150^o$.

\subsection{Properties of DGT-LCC, DGT-DFT and DGT-CCC}\label{subsec:ComProp}
In this subsection, some important properties including unitarity, reversibility and additivity of the DGT-LCC, DGT-DFT and DGT-CCC are discussed.
\\

\noindent\emph{Unitarity property:}

The unitarity property of a DGT is defined as
\begin{align}\label{eq:ComProp06}
{\rm DGT}_0\{g[m,n]\}=g[p,q],
\end{align}
where ${\rm DGT}_\alpha$ denotes the DGT with angle $\alpha$.
Since the DGT-LCC and DGT-DFT have a singularity at $\alpha=0$, we need to make an additional definition that $G[p,q]=g[p,q]$ for $\alpha=0$ just as the continuous gyrator transform does.
When $\alpha=0$, the DGT-CCC reduces to the cascade of a 2D DFT and a 2D IDFT and is equivalent to the identity operator.
Thus, the DGT-CCC itself has the unitarity property.
\\

\noindent\emph{Reversibility property:}

The reversibility property of a DGT is defined as
\begin{align}
g[m,n]&={\rm DGT}_{\alpha}^{-1}{\rm DGT}_{\alpha}\{g[m,n]\}\label{eq:ComProp14_1}\\
&={\rm DGT}_{-\alpha}{\rm DGT}_{\alpha}\{g[m,n]\}.\label{eq:ComProp14_2}
\end{align}
The computational algorithms of the DGT-LCC, DGT-DFT and DGT-CCC are composed of 2D FFTs and pointwise products, all of which are reversible.
Therefore, the inverse transform ${\rm DGT}_{\alpha}^{-1}$  exists for all the three DGTs.
The benefit of ${\rm DGT}_{\alpha}^{-1}={\rm DGT}_{-\alpha}$ is that we don't need to design the inverse DGT additionally.
It can be easily proved that the DGT-DFT and DGT-CCC satisfy (\ref{eq:ComProp14_2}) from their definitions in (\ref{eq:DGT1dft10_1})-(\ref{eq:DGT1dft10_3}) and (\ref{eq:DGT1cc220_1})-(\ref{eq:DGT1cc220_5}).
However, the DGT-LCC doesn't satisfy (\ref{eq:ComProp14_2}) because of the linear convolution in (\ref{eq:DGT1cc110_2}).
The division by the 2D FFT of $\exp \left[ {\frac{j}{2}({m^2}{\Delta _v}{\Delta _x} + {n^2}{\Delta _u}{\Delta _y})\csc \alpha } \right]$ used in the linear deconvolution of ${\rm DGT}_{\alpha}^{-1}$ is not equal to the multiplication by the 2D FFT of $\exp \left[ {\frac{j}{2}({m^2}{\Delta _v}{\Delta _x} + {n^2}{\Delta _u}{\Delta _y})\csc (-\alpha) } \right]$ used in the linear convolution of ${\rm DGT}_{-\alpha}$.
\\

\begin{figure}[t]
\centering
\includegraphics[width=.95\columnwidth,clip=true]{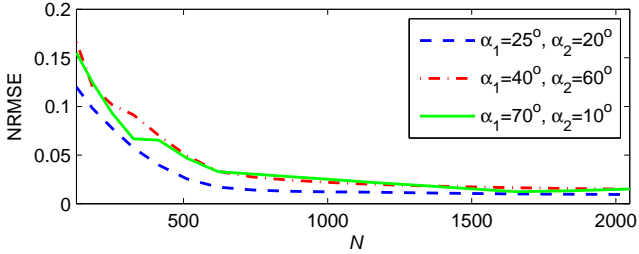}
\caption{
Approximate additivity of DGT-CCC for $(\alpha_1,\alpha_2)=(25^o,20^o)$, $(40^o,60^o)$ and $(70^o,10^o)$: normalized root-mean-square error (NRMSE) (defined in (\ref{eq:ComProp18})) between ${\rm DGT}_{\alpha_2}{\rm DGT}_{\alpha_1}$ and ${\rm DGT}_{\alpha_1+\alpha_2}$ versus the size of the upsampled and zero-padded image, $N\times N$. (The original input is $128\times128$ with  $\Delta_x=\Delta_y=0.1567$.)
}
\label{fig:Add}
\end{figure}

\noindent\emph{Additivity property:}

The additivity property of a DGT is defined as
\begin{align}\label{eq:ComProp08}
{\rm DGT}_{\alpha_2}{\rm DGT}_{\alpha_1}={\rm DGT}_{\alpha_1+\alpha_2}.
\end{align}
The DGT-LCC doesn't satisfy the additivity property because only the central portion of the output is correct 
(refer to Sec.~\ref{subsec:DGT1cc1}).
The DGT-DFT is not additive either because of its constraints on the sampling intervals.
With the 
same input sampling intervals, ${\rm DGT}_{\alpha_2}{\rm DGT}_{\alpha_1}$ and ${\rm DGT}_{\alpha_1+\alpha_2}$ have different output sampling intervals, and thus the outputs are apparently different.
For the DGT-CCC, a simulation is given to examine its additivity.
First, define normalized root-mean-square error (NRMSE) between $g[m,n]$ and $h[m,n]$ as
\begin{align}\label{eq:ComProp18}
{\rm NRMSE} = \frac{{\sqrt {\sum\limits_m^{} {\sum\limits_n^{} {{{\left| {g[m,n] - h[m,n]} \right|}^2}} } } }}{{\sqrt {\sum\limits_m^{} {\sum\limits_n^{} {{{\left| {g[m,n]} \right|}^2}} } } }}.
\end{align}
Given a $128\times128$ Lena image with  $\Delta_x=\Delta_y=0.1567$ as the input,
the NRMSE between ${\rm DGT}_{\alpha_2}{\rm DGT}_{\alpha_1}$ and ${\rm DGT}_{\alpha_1+\alpha_2}$ is $0.1198$ for $(\alpha_1,\alpha_2)=(25^o,20^o)$, $0.1661$ for $(40^o,60^o)$ and $0.1552$ for $(70^o,10^o)$.
Therefore, the DGT-CCC is not additive.
However, as $\Delta_x$ and $\Delta_y$ are reduced by upsampling and more zeros are padded on all sides of the input, the DGT-CCC can approach the continuous gyrator transform. 
And it is expected that the NRMSE will decrease because the continuous gyrator transform has perfect additivity property.
Fig.~\ref{fig:Add} shows the NRMSE versus the size of the upsampled and zero-padded input image, $N\times N$ from $N=128$ (original) to $N=2048$.
It is shown that the DGT-CCC is ``approximate'' additive when $N$ is large enough.

\begin{figure}[t]
\centering
\includegraphics[width=.95\columnwidth,clip=true]{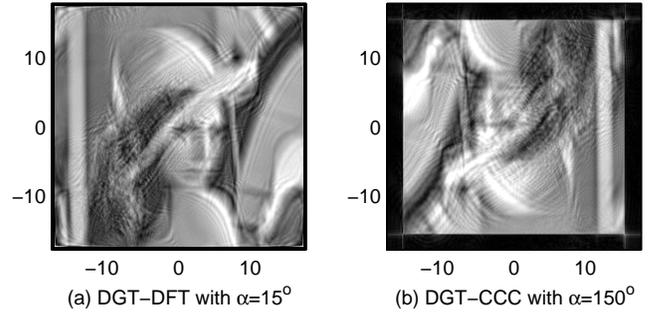}
\caption{
Magnitudes of the DGTs of a $256\times256$ Lena image with ${\Delta _x}={\Delta _y}= 0.14$: (a) $15^o$ DGT-DFT replaced by the cascade of a 2D DFT, a reflection and a $-75^o$ DGT-DFT (see (\ref{eq:DGT208})), and (b) $150^o$ DGT-CCC replaced by  $-30^o$ DGT-CCC of $g[-m,-n]$ (see (\ref{eq:DGT216})).
}
\label{fig:DGT2}
\end{figure}

\subsection{Discrete Gyrator Transforms for $\alpha$ Close to
$k\pi$}\label{subsec:DGT2}
It has been indicated that when $\alpha$ is close to
$k\pi$ or $(2k+1)\pi$, the DGT-LCC, DGT-DFT and DGT-CCC suffer from:
\begin{itemize}
\item[1.] Accuracy
decreases because the absolute values of $\cot\alpha$, $\csc\alpha$ and $\tan\frac{\alpha}{2}$ are too large to be accurately described.
\item[2.] Overlapping (aliasing) effect is produced when the input sampling intervals are not small enough.
\end{itemize}
The upsampling method used in Figs.~\ref{fig:DGT1cc1}, \ref{fig:DGT1dft} and \ref{fig:DGT1cc2} can solve the second problem.
However, as $\alpha$ is much close to
$k\pi$ or $(2k+1)\pi$, very high upsampling rate followed by very high computational complexity is required.
Besides, upsampling cannot solve the accuracy
decreasing problem.

For the DGT-LCC and DGT-DFT, another solution is based on the parameter matrix decomposition below:
\begin{align*}
\mathbf{M}_{\alpha} =
\begin{bmatrix}
{\sin \alpha }&0&0&{-\cos \alpha }\\
0&{\sin \alpha }&{-\cos\alpha }&0\\
0&{  \cos \alpha }&{\sin \alpha }&0\\
{  \cos \alpha }&0&0&{\sin \alpha }
\end{bmatrix}
\begin{bmatrix}
0&1&0&0\\
1&0&0&0\\
0&0&0&1\\
0&0&1&0
\end{bmatrix}
\begin{bmatrix}
0&0&1&0\\
0&0&0&1\\
{ - 1}&0&0&0\\
0&{ - 1}&0&0
 \end{bmatrix}.
\end{align*}
It implies the DGT with $\alpha$ can be calculated by the cascade of a 2D DFT, a reflection and a DGT with $\alpha-\pi/2$, i.e.
\begin{align}\label{eq:DGT208}
&G[p,q] = {\rm{DG}}{{\rm{T}}_{\alpha  - \frac{\pi }{2}}}\left\{ {\frac{{{\Delta _x}{\Delta _y}}}{{2\pi }}\sum\limits_m {\sum\limits_n {g[m,n]} } } \right.\nn\\
&\qquad\qquad\qquad\qquad \cdot \exp \left. {\left( { - j\frac{{2\pi n'm}}{{{N_1}}} - j\frac{{2\pi m'n}}{{{N_2}}}} \right)} \right\}.
\end{align}
Even if $\alpha$ is close to 
$k\pi$, through (\ref{eq:DGT208}), the DGT-LCC and DGT-DFT can still be used because $\alpha-\pi/2$ is far from $k\pi$.
But the cost is one more 2D FFT.
Use 
Fig.~\ref{fig:DGT1dft}(a) as an example.
The DGT-DFT with $\alpha=15^o$ can be replaced by the cascade of a 2D DFT, a reflection and a DGT-DFT with $\alpha-\pi/2=-75^o$.
The new result is shown in Fig.~\ref{fig:DGT2}(a).
Note that $\Delta_u$ and $\Delta_v$ change 
into $\Delta_u=|\sin(\alpha-\pi/2)|\Delta_y$ and $\Delta_v=|\sin(\alpha-\pi/2)|\Delta_x$, respectively, because of the additional 2D DFT.

Because the DGT-CCC is singular only at $\alpha=(2k+1)\pi$, a simpler decomposition is used:
\begin{align*}
\mathbf{M}_{\alpha} =
\begin{bmatrix}
{ - \cos \alpha }&0&0&{ - \sin \alpha }\\
0&{ - \cos \alpha }&{ - \sin \alpha }&0\\
0&{\sin \alpha }&{ - \cos \alpha }&0\\
{\sin \alpha }&0&0&{ - \cos \alpha }
 \end{bmatrix}
\begin{bmatrix}
{ - 1}&0&0&0\\
0&{ - 1}&0&0\\
0&0&{ - 1}&0\\
0&0&0&{ - 1}
\end{bmatrix}.
\end{align*}
In the discrete case, the above equation implies
\begin{align}\label{eq:DGT216}
G[p,q] = {\rm{DG}}{{\rm{T}}_{\alpha  - \pi }}\left\{ {g[ - m, - n]} \right\}.
\end{align}
Therefore, if $\alpha\to(2k+1)\pi$, the DGT can be calculated by the DGT-CCC with $\alpha-\pi$, which is far from $(2k+1)\pi$.
For example, the $150^o$ DGT-CCC 
in Fig.~\ref{fig:DGT1cc2}(d) can be replaced by the $-30^o$ DGT-CCC of $g[ - m, - n]$, as shown in Fig.~\ref{fig:DGT2}(b).

\newcommand{\imineq}[2]{\vcenter{\hbox{\includegraphics[height=#2ex,clip=true]{#1}}}}

\section{
Development of Discrete Gyrator Transform Based on Eigenfunctions}\label{sec:DGTeig}
In this section, we 
develop a DGT based on the eigenfunctions of the continuous gyrator transform.
The 1D Hermite Gaussian function (HGF) of order $k$ is defined as
\begin{align}\label{eq:DGTeig02}
HG_{k}(x) = {\left( {\frac{1}{{{2^{k }}k!\sqrt{\pi} }}} \right)^{1/2}}{e^{ - \frac{x^2}{2}}}{H_k}(x),
\end{align}
where $H_k(x)$ is the $k$th-order
physicists' Hermite polynomial.
The 2D HGF of order $(k,l)$ is a separable function defined as
\begin{align}\label{eq:DGTeig04}
HG_{k,l}(x,y) = HG_{k}(x) HG_{l}(y).
\end{align}
The geometric rotation of the 2D HGF through $45^\circ$ counterclockwise, called rotated HGF (RHGF) for short, is given by
\begin{align}\label{eq:DGTeig06}
RHG_{k,l}(x,y) = H{G_{k,l}}\left( {\frac{{x + y}}{{\sqrt 2 }},\frac{{ - x + y}}{{\sqrt 2 }}} \right).
\end{align}
It has been shown in \cite{pei2009properties} that the RHGF of order $(k,l)$ is the eigenfunction of the continuous gyrator transform with eigenvalue ${e^{ - j\alpha (k - l)}}$; that is
\begin{align}\label{eq:DGTeig08}
{\rm{G}}{{\rm{T}}_\alpha }\left\{ {RH{G_{k,l}}(x,y)} \right\} = {e^{ - j\alpha (k - l)}}RH{G_{k,l}}(u,v).
\end{align}
Since the 2D HGFs can form an orthonormal basis, the RHGFs are also orthonormal to each other.
If the input signal $g(x,y)$ can be expanded by the RHGFs with coefficients ${\widehat g_{k,l}}$, i.e.
\begin{align}\label{eq:DGTeig10}
g(x,y) = \sum\limits_{k = 0}^\infty  \sum\limits_{l = 0}^\infty  {{{\widehat g}_{k,l}}RH{G_{k,l}}(x,y)} ,
\end{align}
\begin{align}\label{eq:DGTeig12}
\!\!\!\!\!\textmd{where}\qquad{\widehat g_{k,l}} =\int\limits_{ - \infty }^\infty \!\int\limits_{ - \infty }^\infty  \! {g(x,y)RH{G_{k,l}}(x,y)dxdy},\quad\
\end{align}
then the gyrator transform can be obtained from
\begin{align}\label{eq:DGTeig14}
G(u,v) = \sum\limits_{k = 0}^\infty  \sum\limits_{l = 0}^\infty   {{e^{ - j\alpha (k - l)}}{{\widehat g}_{k,l}}} RH{G_{k,l}}(u,v).
\end{align}
For the discrete case, if the DGT is obtained by directly sampling (\ref{eq:DGTeig12}) and (\ref{eq:DGTeig14}), it is close to the continuous gyrator transform, but the unitarity, reversibility and additivity properties don't hold anymore because the samples of the RHGFs (sampled RHGFs) cannot form an orthogonal basis.
And it follows that there is no superiority over the 
DGT-LCC, DGT-DFT and DGT-CCC.
In order to retain these important properties, the priority is to generate discrete 
orthonormal RHGFs that approximate the sampled RHGFs.

\subsection{DGT Based On Discrete HGFs (DGT-DHGF)}\label{subsec:DGTdhgf}
It is difficult to directly develop the discrete orthonormal versions of the 2D nonseparable functions, RHGFs. Fortunately, according to \cite{beijersbergen1993astigmatic,varshalovich1988quantum}, there is a relation between the RHGFs and the separable functions, 2D HGFs:
\begin{align}\label{eq:DGTeig16}
RHG_{k,l}(x,y)
= \sum\limits_{s = 0}^L {d_{\frac{{l - k}}{2},\frac{L}{2} - s}^{\frac{L}{2}}\left( { \frac{\pi }{2}} \right)} H{G_{s,L - s}}(x,y),
\end{align}
where $L=k+l$, and $d^J_{M_1M_2}(\beta)$ is the Wigner d-function \cite{varshalovich1988quantum}.
For example, when $L=2$, the $RHG_{0,2},RHG_{1,1},RHG_{2,0}$ are the linear combinations of the $HG_{0,2},HG_{1,1},HG_{2,0}$:
\begin{align}\label{eq:DGTeig17}
\renewcommand{\arraystretch}{2}
\begin{bmatrix}
RHG_{0,2}\, \imineq{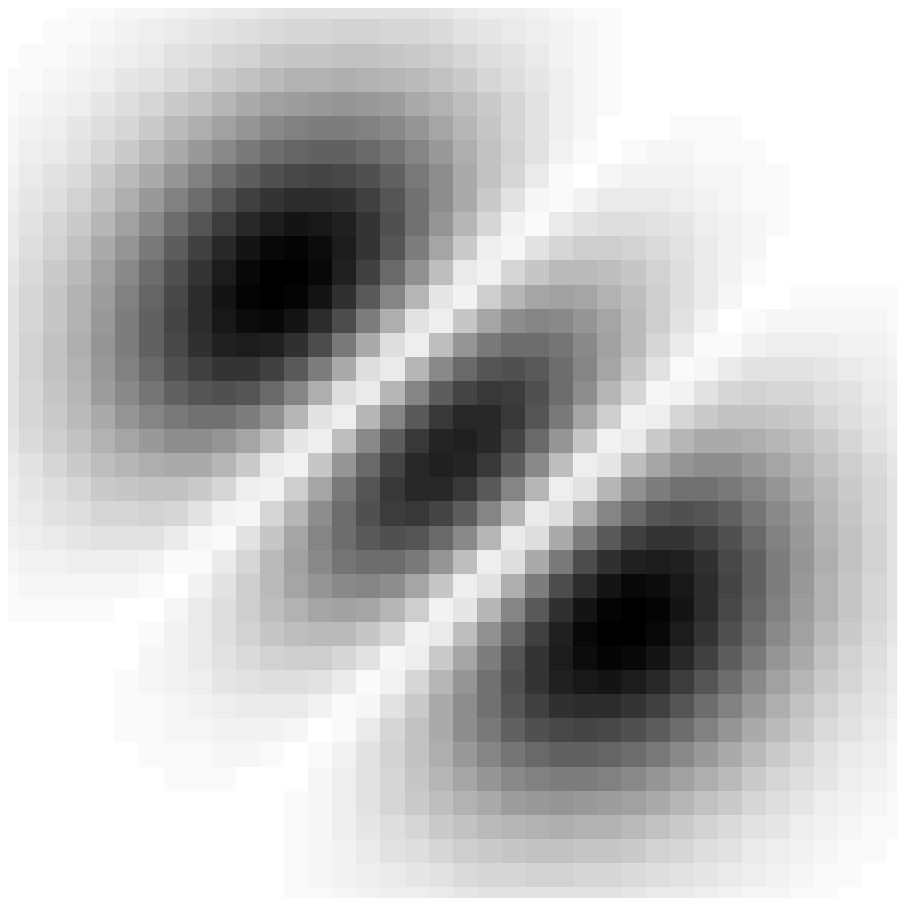}{5}\\
RHG_{1,1}\, \imineq{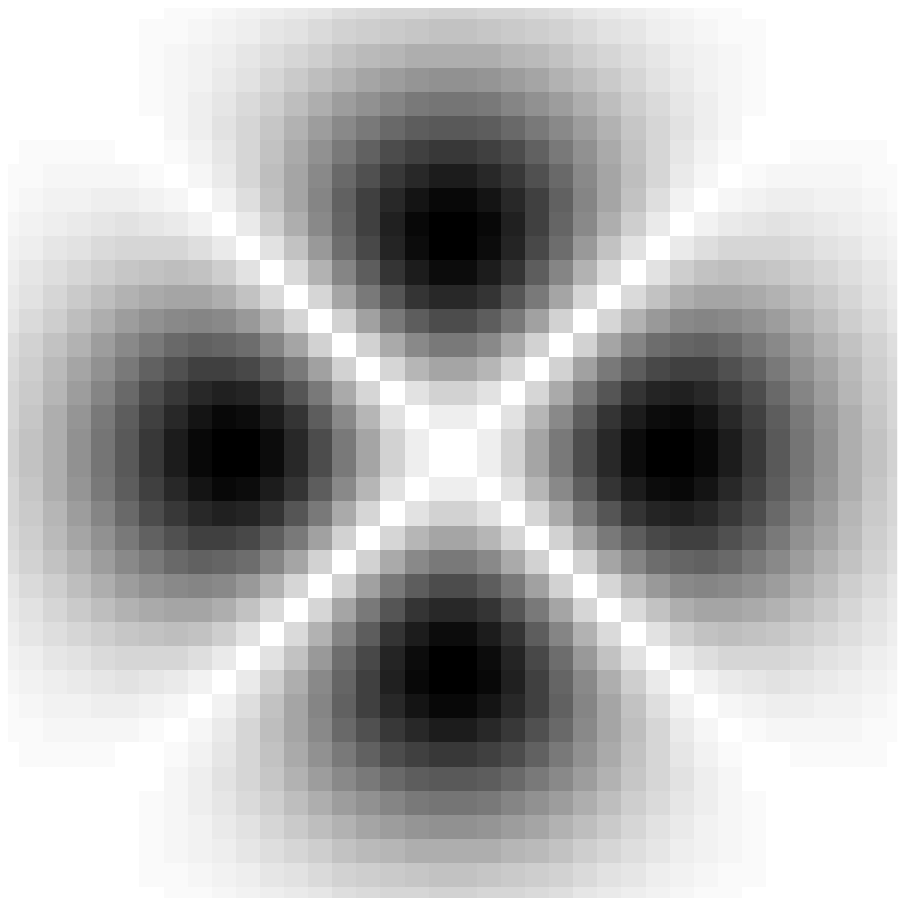}{5}\\
RHG_{2,0}\, \imineq{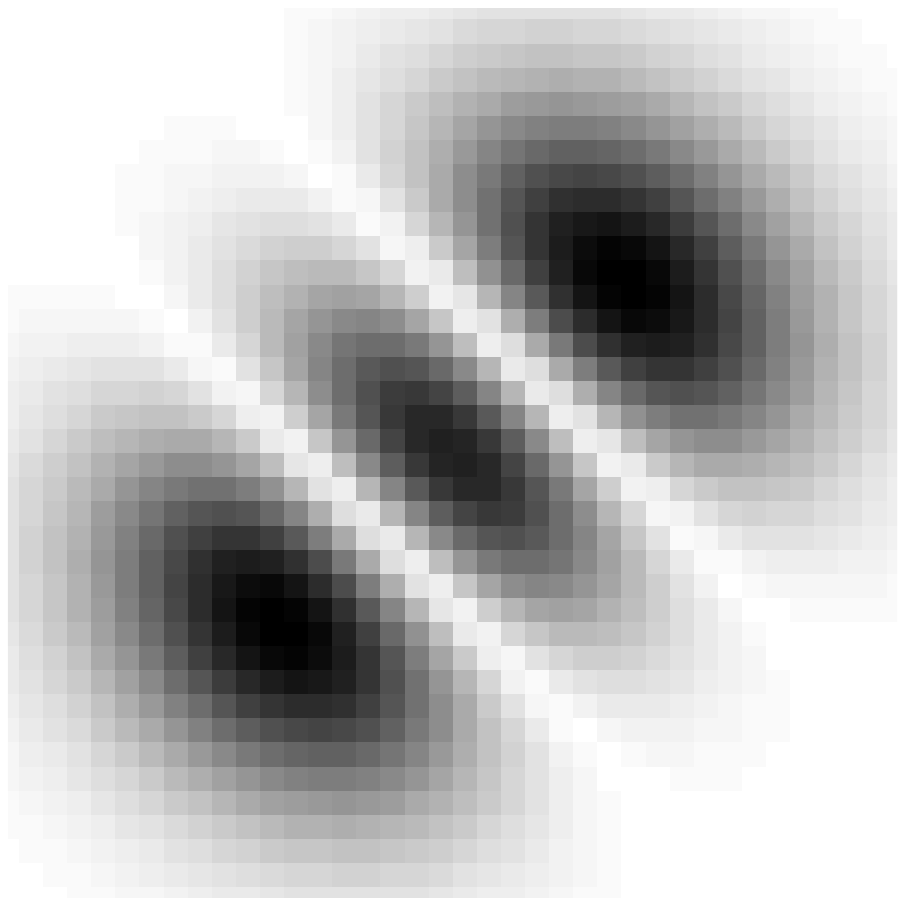}{5}
\end{bmatrix}
\!= \!\begin{bmatrix}
{\frac{1}{2}}&{ - \frac{1}{{\sqrt 2 }}}&{\frac{1}{2}}\\
{\frac{1}{{\sqrt 2 }}}&0&{ - \frac{1}{{\sqrt 2 }}}\\
{\frac{1}{2}}&{\frac{1}{{\sqrt 2 }}}&{\frac{1}{2}}
\end{bmatrix}
\!\begin{bmatrix}
HG_{0,2}\, \imineq{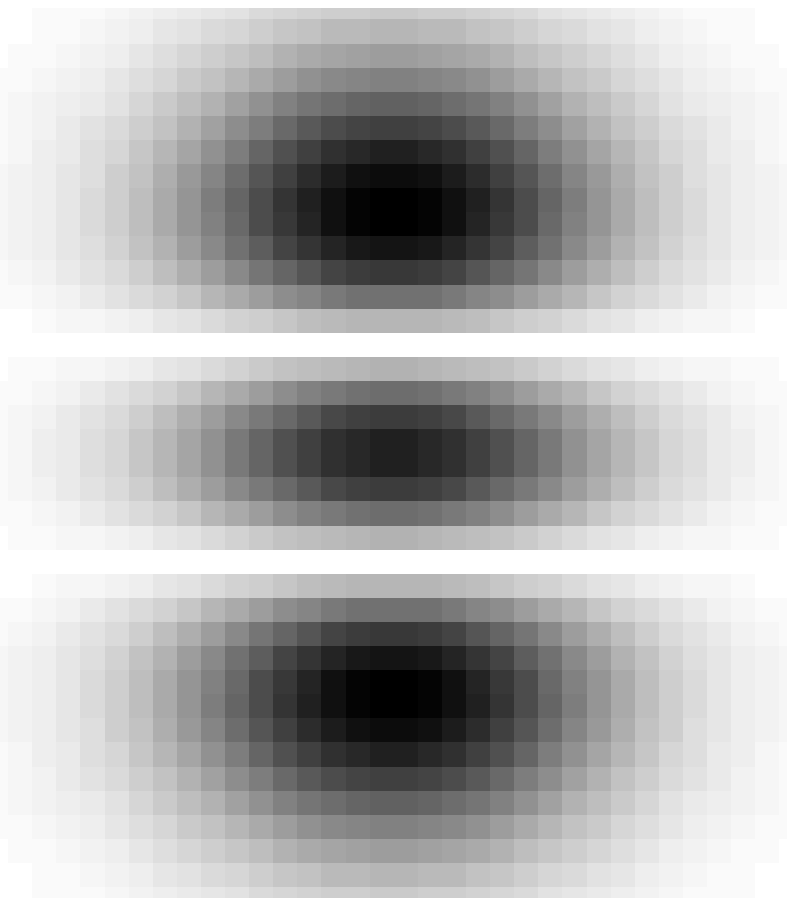}{5}\\
HG_{1,1}\, \imineq{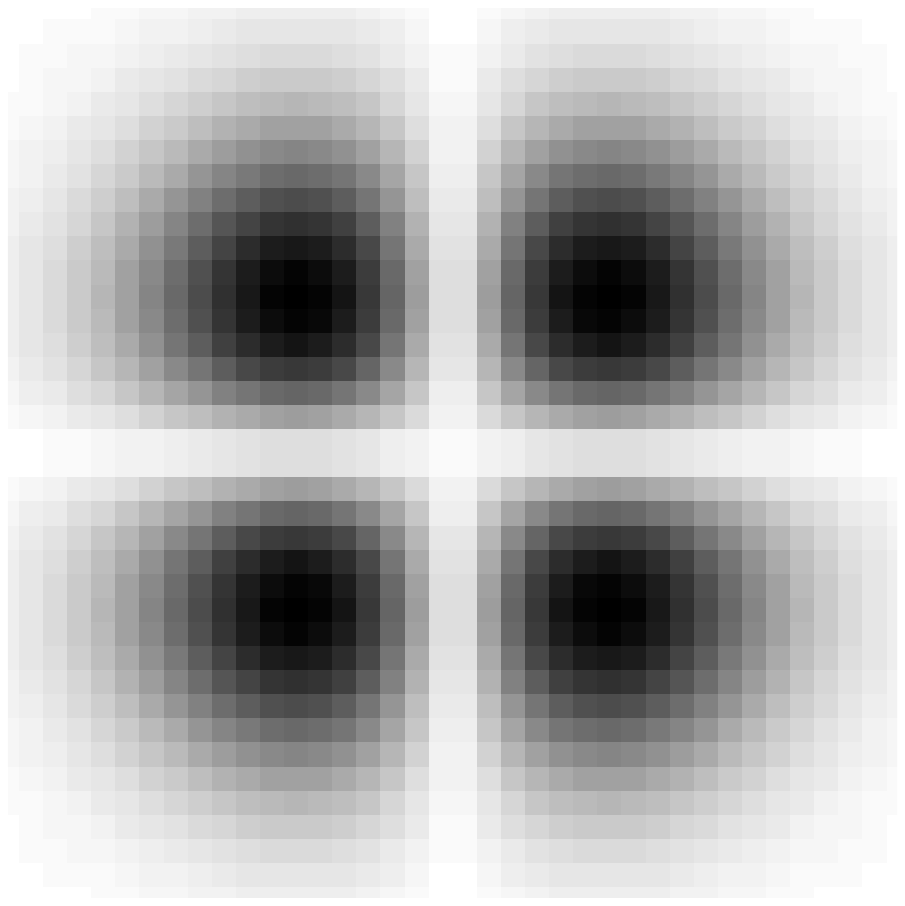}{5}\\
HG_{2,0}\, \imineq{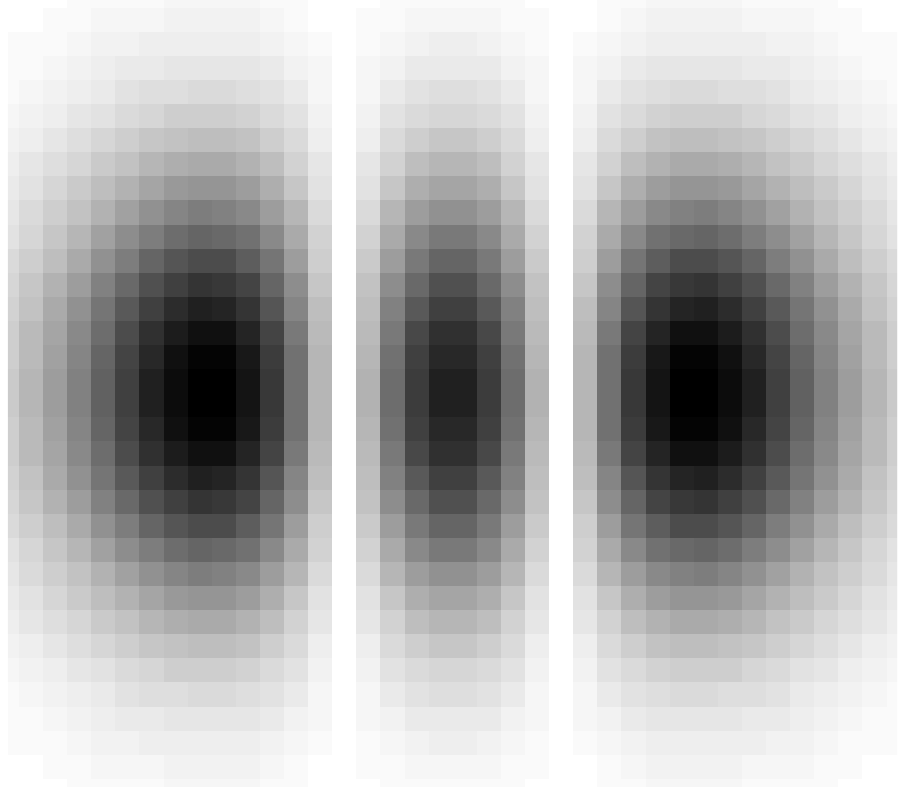}{5}
\end{bmatrix}.
\end{align}
The derivation of (\ref{eq:DGTeig16}) is presented in Appendix \ref{App:RHGF}.
From (\ref{eq:DGTeig16}), the problem is reduced to the development of 1D discrete orthonormal HGFs with good approximation to the samples of the continuous HGFs (sampled HGFs).

\begin{figure}[t]
\centering
\includegraphics[width=.9\columnwidth,clip=true]{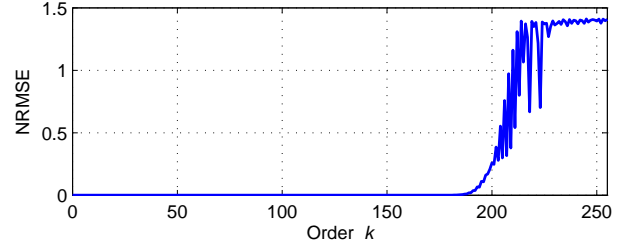}
\caption{
Normalized root-mean-square error (NRMSE) between the discrete HGFs and the sampled HGFs for $N=256$ and orders $k=0,1,\ldots,255$.
}
\label{fig:HGF_RMSE}
\end{figure}

1D discrete orthonormal HGFs have been investigated in numerous studies and are usually generated by the commuting matrices of the DFT \cite{martucci1994symmetric,candan2000discrete,pei2006discrete,santhanam2007discrete,candan2007higher,pei2008generalized}.
Here, the discrete HGFs generated by the so called offset-$n^2$ matrix \cite{pei2008generalized} are adopted.
These discrete HGFs are the orthonormal eigenvectors of the offset-$n^2$ matrix.
Assume the number of discrete points is $N$.
Denote $HG_k[m]$ as the $k$th-order discrete HGF where $0\leq k\leq N-1$.
When $N$ is large enough, $HG_k[m]$ can approximate the 
sampled HGF:
\begin{align}\label{eq:DGTdhgf04}
HG_k[m] \approx c_0 HG_k\left(\left( {m - \frac{{N - 1}}{2}} \right) \sqrt\frac{{2\pi }}{N}\right),
\end{align}
where $0\leq m\leq N-1$, and $c_0$ is used to normalize the sampled HGFs.
The NRMSE between the discrete HGFs and the 
sampled HGFs for $N=256$ is shown in Fig.~\ref{fig:HGF_RMSE}.
If the continuous HGF has energy more concentrate within $\left[ { - \frac{{N - 1}}{2}\sqrt {\frac{{2\pi }}{N}} ,{\rm{ }}\frac{{N - 1}}{2}\sqrt {\frac{{2\pi }}{N}} } \right]$, the corresponding discrete HGF can approximate the sampled HGF with higher accuracy.
Thus, it is inevitable that high-order discrete HGFs are less accurate because they have energy scattered in larger time interval.

If the 2D discrete HGFs are defined as
\begin{align}\label{eq:DGTdhgf06}
HG_{k,l}[m,n] = H{G_k}[m]H{G_l}[n],
\end{align}
the discrete RHGFs can be obtained from the relation in (\ref{eq:DGTeig16}):
\begin{align}\label{eq:DGTdhgf08}
RHG_{k,l}[m,n]
= \sum\limits_{s = 0}^L {d_{\frac{{l - k}}{2},\frac{L}{2} - s}^{\frac{L}{2}}\left( { \frac{\pi }{2}} \right)} HG_{s,L - s}[m,n],
\end{align}
where $L=k+l$.
If the input is of size $N\times N$, there are 
$N^2$ orthonormal 2D discrete HGFs, i.e. $HG_{k,l}$ for $0\leq k,l\leq N-1$.
However, from (\ref{eq:DGTdhgf08}), the calculation of $RHG_{k,l}$ with $k+l=L\geq N$ requires the $HG_{k,l}$ with $k\geq N$ or $l\geq N$.
For example, consider $N=4$.
One can generate an orthonormal set of $16$ 2D discrete HGFs, i.e. $HG_{k,l}$ for $0\leq k,l\leq 3$.
To obtain $RHG_{1,3}$, one requires $HG_{0,4}$, $HG_{1,3}$, $HG_{2,2}$, $HG_{3,1}$ and $HG_{4,0}$; however, $HG_{0,4}$ and $HG_{4,0}$ are not included in the orthonormal set.
To solve this problem, two methods have been proposed in \cite{liu2012discrete}.
In order to let the discrete RHGFs remain orthonormal, the second method ``mirroring the coefficients'' is employed. When $k+l=L\geq N$, (\ref{eq:DGTdhgf08}) is replaced by the following equation:
\begin{align}\label{eq:DGTdhgf10}
RH{G_{k,l}}[m,n] = \!\!\!\!\sum\limits_{s = L - N + 1}^{N - 1}\!\!\! {d_{\frac{{l - k}}{2},N - 1 - \frac{L}{2} - s}^{N - 1 - \frac{L}{2}}\left( {\frac{\pi }{2}} \right)} H{G_{s,L - s}}[m,n].
\end{align}
The above approximation will reduce the accuracy of the high-order discrete RHGFs.

After the $N^2$ discrete RHGFs are obtained, the discrete versions of (\ref{eq:DGTeig12}) and (\ref{eq:DGTeig14}) are given by
\begin{align}
{\widehat g_{k,l}} &= \sum\limits_{m = 0}^{N - 1} {\sum\limits_{n = 0}^{N - 1} {g[m,n]RH{G_{k,l}}[m,n]} },\label{eq:DGTdhgf14}\\
G[p,q] &= \sum\limits_{k = 0}^{N - 1} {\sum\limits_{l = 0}^{N - 1} {{e^{ - j\alpha (k - l)}}{{\widehat g}_{k,l}}} } RH{G_{k,l}}[p,q].\label{eq:DGTdhgf12}
\end{align}
This DGT is basically based on the discrete HGFs, and thus called DGT-DHGF for short.
Note that the input and output sampling intervals are both $\sqrt{2\pi/N}$ because it is used when generating the 1D discrete HGFs (see (\ref{eq:DGTdhgf04})).

\subsection{Characteristics and Properties of DGT-DHGF}\label{subsec:ComProp3}
\begin{figure}[t]
\centering
\includegraphics[width=.95\columnwidth,clip=true]{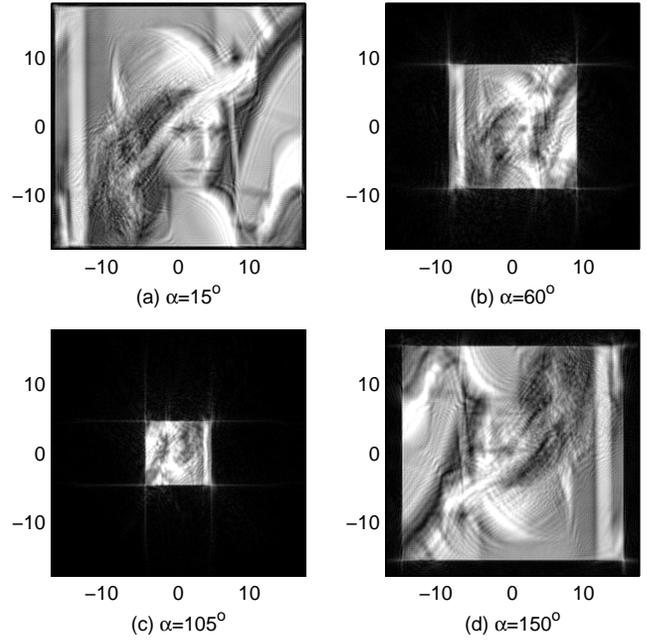}
\caption{
Magnitudes of the DGT-DHGFs of the $320\times320$ zero-padded Lena image (original size is $256\times256$): (a) $\alpha=15^o$, (b) $\alpha=60^o$, (c) $\alpha=105^o$, and (d) $\alpha=150^o$, where ${\Delta _x}={\Delta _y}={\Delta _u}={\Delta _v}=0.14$.
(Only the central $256\times256$ data of the outputs are displayed.)
}
\label{fig:DGT_DHGF}
\end{figure}

It is apparent that the DGT-DHGF is suitable for all angles.
A simulation of 
the DGT-DHGFs of the $256\times256$ Lena image with $\alpha$ being $15^o$, $60^o$ $105^o$ and $150^o$ is given.
Because of the less accurate high-order discrete HGFs (see Fig.~\ref{fig:HGF_RMSE}) and the approximation in (\ref{eq:DGTdhgf10}), high-order discrete RHGFs have much lower accuracy than the low-order ones.
This will yield higher error at the boundary of the output of the DGT-DHGF.
A simple solution for this problem is zero-padding the input signal/image.
In this simulation, the input image is 
zero-padded to $320\times320$.
The central $256\times256$  output data of the DGT-DHGFs are shown in Fig.~\ref{fig:DGT_DHGF}.
The sampling intervals are $\Delta_x=\Delta_y=\Delta_u=\Delta_v=\sqrt{2\pi/320}=0.14$.
Unlike the DGT-LCC, DGT-DFT and DGT-CCC, the DGT-DHGF doesn't have overlapping
(aliasing) problem.
\\

\noindent\emph{Unitarity property:}

Denote the DGT-DHGF as ${\rm DGT}^{\rm DHGF}_\alpha$.
Since $RH{G_{k,l}}$'s form an orthonormal set, the unitarity property can be easily proved by setting $\alpha=0$ in (\ref{eq:DGTdhgf12}). That is,
\begin{align}\label{eq:DGTdhgf16}
{\rm{DGT}}_0^{{\rm{DHGF}}}\left\{ {g[m,n]} \right\} &= \sum\limits_{k = 0}^{N - 1} {\sum\limits_{l = 0}^{N - 1} {{{\widehat g}_{k,l}}} } RH{G_{k,l}}[p,q]\nn\\
&=g[p,q].
\end{align}

\noindent\emph{Additivity property:}

For the additivity property, we want to prove
\begin{align}\label{eq:DGTdhgf18}
{\rm DGT}^{\rm DHGF}_{\alpha_2}{\rm DGT}^{\rm DHGF}_{\alpha_1}={\rm DGT}^{\rm DHGF}_{{\alpha_2}+{\alpha_1}}.
\end{align}
The $G[p,q]={\rm{DGT}}_{\alpha_1}^{{\rm{DHGF}}}\left\{ {g[m,n]} \right\}$ is given in (\ref{eq:DGTdhgf14}) and (\ref{eq:DGTdhgf12})
when $\alpha=\alpha_1$. Therefore, $G'[p',q']={\rm{DGT}}_{\alpha_2}^{{\rm{DHGF}}}\left\{ {G[p,q]} \right\}$ can be obtained from
\begin{align}\label{eq:DGTdhgf20}
{{\widehat G}_{k,l}} = \!\sum\limits_{p = 0}^{N - 1} {\sum\limits_{q = 0}^{N - 1} {G[p,q]RH{G_{k,l}}[p,q]} }  = {e^{ - j{\alpha_1} (k - l)}}{{\widehat g}_{k,l}},
\end{align}
\begin{align}\label{eq:DGTdhgf22}
G'[p',q'] &= \sum\limits_{k = 0}^{N - 1} {\sum\limits_{l = 0}^{N - 1} {{e^{ - j{\alpha_2} (k - l)}}{{\widehat G}_{k,l}}} } RH{G_{k,l}}[p',q']\nn\\
 &= \sum\limits_{k = 0}^{N - 1} {\sum\limits_{l = 0}^{N - 1} {{e^{ - j({\alpha_2}  + {\alpha_1} )(k - l)}}{{\widehat g}_{k,l}}} } RH{G_{k,l}}[p',q']\nn\\
 &= {\rm{DGT}}_{{\alpha_2}  + {\alpha_1} }^{{\rm{DHGF}}}\left\{ {g[m,n]} \right\}.
\end{align}

\noindent\emph{Reversibility property:}

This property can be easily proved from the unitary property in (\ref{eq:DGTdhgf16}) and additivity property in (\ref{eq:DGTdhgf18}) with $\alpha_2=-\alpha_1$:
\begin{align}\label{eq:DGTdhgf24}
{\rm{DGT}}_{ - {\alpha _1}}^{{\rm{DHGF}}}{\rm{DGT}}_{{\alpha _1}}^{{\rm{DHGF}}}\left\{ {g[m,n]} \right\} &= {\rm{DGT}}_0^{{\rm{DHGF}}}\left\{ {g[m,n]} \right\}\nn\\
&= g[p,q].
\end{align}

\subsection{Efficient Computational Algorithm for DGT-DHGF}\label{subsec:AlgDHGF}
Substituting (\ref{eq:DGTeig12}) and (\ref{eq:DGTeig16}) into (\ref{eq:DGTeig14}), the gyrator transform can be expressed in terms of 2D HGFs, i.e.
\begin{align}\label{eq:DGTeig18}
G(u,v) &= \sum\limits_{k' = 0}^\infty  {\sum\limits_{l' = 0}^\infty  \left[{\sum\limits_{r = 0}^L {D_{\frac{{l' - k'}}{2},\frac{L}{2} - r}^{\frac{L}{2}}\left( { - \frac{\pi }{2},2\alpha ,\frac{\pi }{2}} \right)} } {{\widetilde g}_{r,L-r}} \right]} \nn\\
&\qquad\qquad\qquad\qquad\qquad \qquad\cdot H{G_{k',l'}}(u,v),
\end{align}
\begin{align}\label{eq:DGTeig20}
\!\!\!\!\!\textmd{where}\qquad{{\widetilde g}_{k,l}} = \int\limits_{ - \infty }^\infty   \int\limits_{ - \infty }^\infty   g(x,y)H{G_{k,l}}(x,y)dxdy,\quad\ \
\end{align}
and $D_{{M_1},{M_2}}^J\left( {\chi ,\beta ,\gamma } \right)$ denotes the Wigner D-function \cite{varshalovich1988quantum} defined as
\begin{align}\label{eq:DGTeig22}
D_{{M_1},{M_2}}^J\left( {\chi ,\beta ,\gamma } \right) = {e^{-j{M_1}\chi }}d_{{M_1},{M_2}}^J\left( \beta  \right){e^{-j{M_2}\gamma }}.
\end{align}
The detailed derivation of (\ref{eq:DGTeig18}) is available in Appendix \ref{App:RHGF}.
From (\ref{eq:DGTeig18}) and (\ref{eq:DGTeig20}), the DGT-DHGF can also be calculated by the following three steps:
\begin{align}
{{\widetilde g}_{k,l}} &= \sum\limits_{m = 0}^{N - 1} {\sum\limits_{n = 0}^{N - 1} {g[m,n]H{G_{k,l}}[m,n]} } ,\label{eq:DGTdhgf15_1}\\
{{\widetilde G}_{k',l'}} &= \sum\limits_{r = 0}^L  D_{\frac{{l' - k'}}{2},\frac{L}{2} - r}^{\frac{L}{2}}\left( { - \frac{\pi }{2},2\alpha ,\frac{\pi }{2}} \right){{{\widetilde g}_{r,L - r}}},\label{eq:DGTdhgf15_2}\\
G[p,q] &= \sum\limits_{k' = 0}^{N - 1} {\sum\limits_{l' = 0}^{N - 1} {{{\widetilde G}_{k',l'}}} } H{G_{k',l'}}[p,q],\label{eq:DGTdhgf15_3}
\end{align}
where $L=k'+l'$.
As $L\geq N$, the second step in (\ref{eq:DGTdhgf15_2}) suffers from the same problem encountered in the generation of discrete RHGFs in (\ref{eq:DGTdhgf08}).
Therefore, the method of mirroring the coefficients used in  (\ref{eq:DGTdhgf10}) is applied to (\ref{eq:DGTdhgf15_2})  when $L\geq N$:
\begin{align}\label{eq:DGTdhgf15_4}
{{\widetilde G}_{k',l'}}\! = \!\!\!\sum\limits_{r=L-N+1}^{N-1}
\!\!
 D_{\frac{{l' - k'}}{2},N - 1 - \frac{L}{2} - r}^{N - 1 - \frac{L}{2}}
\!\! \left( { - \frac{\pi }{2},2\alpha ,\frac{\pi }{2}} \right)
 {{{\widetilde g}_{r,L - r}}}.
\end{align}

Compared with (\ref{eq:DGTdhgf14})-(\ref{eq:DGTdhgf12}), the computational algorithm (\ref{eq:DGTdhgf15_1})-(\ref{eq:DGTdhgf15_4}) is much more efficient.
The discrete RHGFs are nonseparable, and thus the  $N\times N$ pointwise products in  (\ref{eq:DGTdhgf14}) and (\ref{eq:DGTdhgf12}) need to be performed $N^2$ times for $0\leq k,l\leq N-1$ and $0\leq p,q\leq N-1$, respectively.
On the contrary, since the 2D discrete HGFs are separable, (\ref{eq:DGTdhgf15_1}) and (\ref{eq:DGTdhgf15_3}) can be realized  by four $N\times N$ matrix multiplications.
We give a simple example that $N=3$ to explain the matrix forms of (\ref{eq:DGTdhgf15_1})-(\ref{eq:DGTdhgf15_4}).
Assume ${{\bf{H}}}$ is an $N\times N$ matrix composed of 1D discrete HGFs:
\setlength\arraycolsep{0.4em}\renewcommand{\arraystretch}{1.2}
\begin{align}\label{eq:DGTdhgf36}
\bb H=
\begin{bmatrix}
{H{G_0}[0]}&{H{G_1}[0]}&{H{G_2}[0]}\\
{H{G_0}[1]}&{H{G_1}[1]}&{H{G_2}[1]}\\
{H{G_0}[2]}&{H{G_1}[2]}&{H{G_2}[2]}
\end{bmatrix}.
\end{align}
The matrix form of (\ref{eq:DGTdhgf15_1}) is given by
\begin{align}\label{eq:DGTdhgf32}
\begin{bmatrix}
{{{\widetilde g}_{0,0}}}&{{{\widetilde g}_{1,0}}}&{{{\widetilde g}_{2,0}}}\\
{{{\widetilde g}_{0,1}}}&{{{\widetilde g}_{1,1}}}&{{{\widetilde g}_{2,1}}}\\
{{{\widetilde g}_{0,2}}}&{{{\widetilde g}_{1,2}}}&{{{\widetilde g}_{2,2}}}
\end{bmatrix}={{\bf{H}}^T}
\begin{bmatrix}
{g[0,0]}&{g[1,0]}&{g[2,0]}\\
{g[0,1]}&{g[1,1]}&{g[2,1]}\\
{g[0,2]}&{g[1,2]}&{g[2,2]}
\end{bmatrix}{{\bf{H}}}.
\end{align}
For $L<N=3$, (\ref{eq:DGTdhgf15_2}) is used with matrix forms given by
\begin{align}
{{\widetilde G}_{0,0}} &= D_{0,0}^0\ {{\widetilde g}_{0,0}} = {{\widetilde g}_{0,0}},\label{eq:DGTdhgf40_0}\\
\begin{bmatrix}
{{{\widetilde G}_{0,1}}}\\
{{{\widetilde G}_{1,0}}}
\end{bmatrix}&=
\begin{bmatrix}
{D_{1/2,1/2}^{1/2}}&{D_{1/2, - 1/2}^{1/2}}\\
{D_{ - 1/2,1/2}^{1/2}}&{D_{ - 1/2, - 1/2}^{1/2}}
\end{bmatrix}
\begin{bmatrix}
{{{\widetilde g}_{0,1}}}\\
{{{\widetilde g}_{1,0}}}
\end{bmatrix},\label{eq:DGTdhgf40_1}\\
\begin{bmatrix}
{{{\widetilde G}_{0,2}}}\\
{{{\widetilde G}_{1,1}}}\\
{{{\widetilde G}_{2,0}}}
\end{bmatrix}&=
\begin{bmatrix}
{D_{1,1}^1}&{D_{1,0}^1}&{D_{1, - 1}^1}\\
{D_{0,1}^1}&{D_{0,0}^1}&{D_{0, - 1}^1}\\
{D_{ - 1,1}^1}&{D_{ - 1,0}^1}&{D_{ - 1, - 1}^1}
\end{bmatrix}
\begin{bmatrix}
{{{\widetilde g}_{0,2}}}\\
{{{\widetilde g}_{1,1}}}\\
{{{\widetilde g}_{2,0}}}
\end{bmatrix},\label{eq:DGTdhgf40_2}
\end{align}
where the arguments $\left( { - \frac{\pi }{2},2\alpha ,\frac{\pi }{2}}\right)$ are omitted for brevity.
For $L\geq 3$, (\ref{eq:DGTdhgf15_2}) is replaced by (\ref{eq:DGTdhgf15_4}), and the matrix forms are
\begin{align}
\setlength\arraycolsep{0.4em}\renewcommand{\arraystretch}{1.2}
\begin{bmatrix}
{{{\widetilde G}_{1,2}}}\\
{{{\widetilde G}_{2,1}}}
\end{bmatrix}&=
\begin{bmatrix}
{D_{1/2,1/2}^{1/2}}&{D_{1/2, - 1/2}^{1/2}}\\
{D_{ - 1/2,1/2}^{1/2}}&{D_{ - 1/2, - 1/2}^{1/2}}
\end{bmatrix}
\begin{bmatrix}
{{{\widetilde g}_{1,2}}}\\
{{{\widetilde g}_{2,1}}}
\end{bmatrix},\label{eq:DGTdhgf40_3}\\
{{\widetilde G}_{2,2}} &= D_{0,0}^0\ {{\widetilde g}_{2,2}} = {{\widetilde g}_{2,2}}.\label{eq:DGTdhgf40_4}
\end{align}
At last, (\ref{eq:DGTdhgf15_3}) can be calculated from the following two matrix multiplications:
\setlength\arraycolsep{0.2em}\renewcommand{\arraystretch}{1.2}
\begin{align}\label{eq:DGTdhgf44}
\begin{bmatrix}
{G[0,0]}&{G[1,0]}&{G[2,0]}\\
{G[0,1]}&{G[1,1]}&{G[2,1]}\\
{G[0,2]}&{G[1,2]}&{G[2,2]}
\end{bmatrix}\!={{\bf{H}}}
\begin{bmatrix}
{{{\widetilde G}_{0,0}}}&{{{\widetilde G}_{1,0}}}&{{{\widetilde G}_{2,0}}}\\
{{{\widetilde G}_{0,1}}}&{{{\widetilde G}_{1,1}}}&{{{\widetilde G}_{2,1}}}\\
{{{\widetilde G}_{0,2}}}&{{{\widetilde G}_{1,2}}}&{{{\widetilde G}_{2,2}}}
\end{bmatrix}
{{\bf{H}}^T}.
\end{align}
\setlength\arraycolsep{0.1em}\renewcommand{\arraystretch}{1}

We summarize the computational algorithm of the DGT-DHGF as follows:
\begin{align}
\bb{\widetilde g}&=\bb H^T \bb g \bb H,\label{eq:DGTdhgf48_1}\\
{{{\bf{\widetilde G}}}_L}& = \left\{ {\begin{array}{*{20}{l}}
{{{\bf{D}}_L}\ {{{\bf{\widetilde g}}}_L},}&{\rm{\ \ for\ \ }}{0\leq L \leq N-1}\\
{{{\bf{D}}_{2(N-1) - L}}\ {{{\bf{\widetilde g}}}_L},}&{\rm{\ \ for\ \ }}{N \le L \le 2(N - 1)}
\end{array}} \right.,\label{eq:DGTdhgf48_2}\\
\bb G&=\bb H \widetilde {\bb G}\bb H^T.\label{eq:DGTdhgf48_3}
\end{align}
$\bb g$, $\bb{\widetilde g}$, $\bb G$ and $\bb{\widetilde G}$ are $N\times N$ matrices with the $(i+1,j+1)$-th entry being $g[j,i]$, ${\widetilde g}_{j,i}$, $G[j,i]$ and ${\widetilde G}_{j,i}$, respectively.
$\bb H$ is an $N\times N$ matrix that the $(k+1)$-th column is the $k$-th order 1D discrete HGF.
$\bb{\widetilde g}_{L}$ and $\bb{\widetilde G}_{L}$ are $(L+1)\times1$ vectors with entries ${\widetilde g}_{k,l}$'s and ${\widetilde G}_{k,l}$'s, respectively, where $k+l=L$.
And ${{\bf{D}}_L}$ is an $(L+1)\times (L+1)$ matrix with the $(i+1,j+1)$-th entry being $D_{\frac{L}{2}-i,\frac{L}{2}-j}^{\frac{L}{2}}\left( { - \frac{\pi }{2},2\alpha ,\frac{\pi }{2}}\right)$.
The dominant computational complexity is on (\ref{eq:DGTdhgf48_1}) and (\ref{eq:DGTdhgf48_3}), i.e. four $N\times N$ matrix multiplications.
Taking the benefit of computing the 2D DFT/IDFT by 2D FFT, the complexities of the DGT-LCC, DGT-DFT and DGT-CCC are reduced.
If fast algorithm for (\ref{eq:DGTdhgf48_1}) and (\ref{eq:DGTdhgf48_3}) is developed, the complexity of the DGT-DHGF can further be lowered.

\section{Complexity, Memory and Accuracy}\label{sec:ComAcc}
In this section, we analyze the computational complexity,
memory requirements and accuracy of the proposed DGTs.

\subsection{Complexity}\label{subsec:Complexity}
The complexity of DGTs is measured in terms of number of real multiplications.
Consider that the input is of size $N\times N$.
Directly calculating the $N\times N$ DGT output by summation in (\ref{eq:intro08}) involves $N^4$ complex multiplications, i.e. $4N^4$ real multiplications.
Recall the DGT-LCC in (\ref{eq:DGT1cc110_1})-(\ref{eq:DGT1cc110_3}).
In the first and third steps, the chirp multiplication is implemented by pointwise product of two $N\times N$ matrices, which requires $N^2$ complex multiplications.
In the second step, as mentioned in the second paragraph of Sec.~\ref{subsec:DGT1cc1}, the linear convolution is realized by three FFTs and one pointwise product, all of which are $(3N-2)\times(3N-2)$.
Therefore, the number of real multiplications for DGT-LCC is
\begin{align}\label{eq:ComAcc04}
&4\left[ {2{N^2} + {{(3N - 2)}^2}+ 3 \cdot \frac{{{{(3N - 2)}^2}}}{2}{{\log }_2}{{(3N - 2)}^2}  } \right]\nn\\
& = 8{N^2} + 4{(3N - 2)^2} + 6{(3N - 2)^2}{\log _2}{(3N - 2)^2}.
\end{align}
The DGT-DFT in (\ref{eq:DGT1dft10_1})-(\ref{eq:DGT1dft10_3}) is much simpler, requiring two $N\times N$ pointwise products for the two chirp multiplications and one $N\times N$ FFT for the 2D DFT/IDFT.
It follows that the computational complexity is given by
\begin{align}\label{eq:ComAcc08}
4\left[ {2{N^2} + \frac{{{N^2}}}{2}{{\log }_2}{N^2}} \right] = 8{N^2} + 2{N^2}{\log _2}{N^2}.
\end{align}
The fast algorithm of the DGT-CCC is a composite of three $N\times N$ pointwise products and two $N\times N$ FFTs according to the five steps in (\ref{eq:DGT1cc220_1})-(\ref{eq:DGT1cc220_5}).
That is, the number of real multiplications is
\begin{align}\label{eq:ComAcc12}
4\left[ {3{N^2} + 2 \cdot \frac{{{N^2}}}{2}{{\log }_2}{N^2}} \right] = 12{N^2} + 4{N^2}{\log _2}{N^2}.
\end{align}
For the DGT-DHGF, the $\bb H^T \bb g \bb H$ in (\ref{eq:DGTdhgf48_1}) and $\bb H \widetilde {\bb G}\bb H^T$ in (\ref{eq:DGTdhgf48_3}) are calculated by four matrix-matrix multiplications.
Since $\bb H$ is real, $4N^3\times2=8N^3$ real multiplications are required.
(To our knowledge the fastest known matrix multiplication has an asymptotic complexity of $O(N^{2.3728639})$ \cite{gall2014powers}.)
The second step in (\ref{eq:DGTdhgf48_2}) contains $2N-1$ matrix-vector multiplications with $1^2+2^2+\ldots+N^2+(N-1)^2+\ldots+1^2$ complex multiplications involved.
Accordingly, the total number of real multiplications required in the DGT-DHGF is
\begin{align}\label{eq:ComAcc16}
\!\!\!8{N^3} \!+ 4\left[  2\frac{{(N - 1)N(2N - 1)}}{6} +N^2\right]
\!\!= \!\frac{{32}}{3}{N^3} + \frac{4}{3}N.
\end{align}
We conclude that the order of computational complexity from low to high is
\begin{align}\label{eq:ComAcc18}
&\!\!\textmd{DGT-DFT} {\scriptstyle (O(N^2\!\log\! N))}<\textmd{DGT-CCC} {\scriptstyle (O(N^2\!\log\! N))}\nn\\
&\!<\textmd{DGT-LCC} {\scriptstyle (O(N^2\!\log\! N))}<\textmd{DGT-DHGF}{\scriptstyle (O(N^3))}\nn\\
&\qquad\qquad\qquad\qquad\qquad\ \ <\textmd{Direct summation}{\scriptstyle (O(N^4))},\!\!
\end{align}
but note that the DGT-DHGF would have lower complexity than the DGT-LCC if $N$ is not large enough.

\subsection{Memory}\label{subsec:Memory}
Suppose the input and output are both of size $N\times N$ for simplicity, and adopt $\Delta_x=\Delta_y=\Delta_u=\Delta_v=\sqrt{2\pi/N}$ which is suitable for all the DGTs to make a fair comparison.
The memory requirement of each DGT is closely related to its computational complexity presented in the previous subsection.

As mentioned previously, the direct summation method in (\ref{eq:intro08}) involves $N^4$ complex multiplications.
This is based on the assumption that the exponential kernel function is precomputed for all sampling points $m,n,p,q$.
It implies that $2N^4$ storage registers are required for the $N^4$ complex numbers.
With another $2N^2$ registers shared by the input and output, the memory requirement of the direct summation method is
$2N^4+2N^2$.

Recall the DGT-LCC in (\ref{eq:DGT1cc110_1})-(\ref{eq:DGT1cc110_3}).
In the first step, the $N^2$ complex numbers are precomputed from the exponential term and stored in $2N^2$ registers.
In the second step, three 2D FFTs and one pointwise product are used, all of which are $(3N-2)\times(3N-2)$.
Therefore, $2(3N-2)^2$ more storage registers are used by the $(3N-2)^2$ complex numbers, 2D FFT of ${e^{\frac{j}{2}\left( {{{(q - m)}^2}{\Delta _v}{\Delta _x} + {{(p - n)}^2}{\Delta _u}{\Delta _y}} \right)\csc \alpha }}$.
The memory requirement of the twiddle factors in the 2D FFT can be disregarded as $N$ is large enough.
In the third step, the exponential term is the same as that in the first step because $\Delta_x=\Delta_y=\Delta_u=\Delta_v$.
Thus, no more registers are required.
Since the second step operates on $(3N-2)\times(3N-2)$, we use $2(3N-2)^2$ storage registers for the input, output and intermediate outputs, i.e. $g$, $g_1$, $G_1$ and $G$.
It follows that the memory requirement of the DGT-LCC is about
$2N^2+2\cdot2(3N-2)^2=38{N^2} - 48N + 16$.

For the DGT-DFT in (\ref{eq:DGT1dft10_1})-(\ref{eq:DGT1dft10_3}), the exponential term in the first step is precomputed and stored in $2N^2$ storage registers and can be reused in the third step because $\Delta_x=\Delta_y=\Delta_u=\Delta_v$.
Another $2N^2$ storage registers are shared by the input, output and intermediate outputs.
Accordingly, for the DGT-DFT, the memory requirement is about $4N^2$.

For the DGT-CCC, the three exponential terms in (\ref{eq:DGT1cc220_1}), (\ref{eq:DGT1cc220_3}) and (\ref{eq:DGT1cc220_5}) are precomputed.
The first one and third one are the same when $\Delta_x=\Delta_y=\Delta_u=\Delta_v$.
Therefore, $2\cdot2N^2=4N^2$  storage registers are required.
Plus $2N^2$ storage registers for the input, output and intermediate outputs, the total amount of registers required by the DGT-CCC is about $6N^2$.

Recall the DGT-DHGF in (\ref{eq:DGTdhgf48_1})-(\ref{eq:DGTdhgf48_3}).
The $\bb H$ used in (\ref{eq:DGTdhgf48_1}) and (\ref{eq:DGTdhgf48_3}) and the $\bb D_L$ used in (\ref{eq:DGTdhgf48_2}) are precomputed to reduce the complexity.
Because the $N\times N$ matrix $\bb H$ is real, it requires only $N^2$ storage registers.
The $\bb D_L$'s with $L=0,1,\cdots,2(N-1)$ have $1^2, 2^2,\cdots,N^2, (N-1)^2,\cdots,1^2$ complex elements, respectively, totally requiring $2  \left[ {2\frac{{(N - 1)N(2N - 1)}}{6} + {N^2}} \right] = \frac{4}{3}{N^3} + \frac{2}{3}N$ storage registers.
With another $2N^2$ registers shared by the input, output and intermediate outputs, the memory requirement of the DGT-DHGF is
$\frac{4}{3}{N^3} + 3{N^2} + \frac{2}{3}N$.

Therefore, the order of memory requirements of the DGTs from low to high is also
\begin{align}\label{eq:Mem20}
&\!\!\textmd{DGT-DFT} {\scriptstyle (O(N^2))}<\textmd{DGT-CCC} {\scriptstyle (O(N^2))}<\textmd{DGT-LCC} {\scriptstyle (O(N^2))}\nn\\
&\quad <\textmd{DGT-DHGF}{\scriptstyle (O(N^3))}<\textmd{Direct summation}{\scriptstyle (O(N^4))}.
\end{align}
Note that the DGT-DHGF may require less memory than the DGT-LCC when $N$ is small.

\begin{figure}[t]
\centering
\includegraphics[width=\columnwidth ,clip=true]{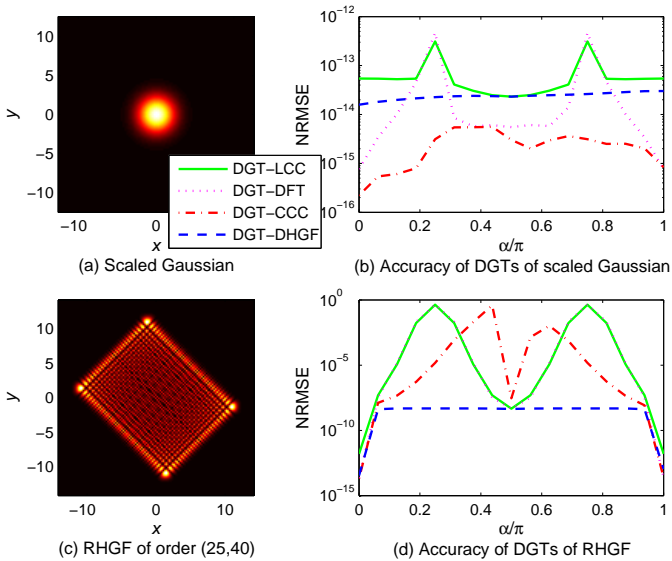}
\caption{
Accuracy of the proposed DGTs for two kinds of inputs:
(a) sampled scaled Gaussian with scaling parameter $s=0.4$, $\Delta_x=\Delta_y=\sqrt{2\pi/N}$ and $N=101$, (b) normalized root-mean-square error (NRMSE) of the DGTs versus $\alpha$ using sampled scaled Gaussian as the input,
(c) sampled RHGF of order $(25,40)$ with $\Delta_x=\Delta_y=\sqrt{2\pi/N}$ and $N=128$, (d) NRMSE of the DGTs versus $\alpha$ using sampled RHGF as the input.
The method in (\ref{eq:DGT208}) or (\ref{eq:DGT216}) is used when $\alpha$ is close to the singularities.
}
\label{fig:Accuracy}
\end{figure}

\begin{table*}
\footnotesize
\begin{center}
\setstretch{1.5}
\caption{Comparisons of the proposed DGTs}\label{tab:table1}
\begin{tabular}{|l|c|c|c|c|}
\hline
 &DGT-LCC & DGT-DFT & DGT-CCC  & DGT-DHGF \\
\hline\hline
Unitarity (${\rm DGT}_{0}$  exists)& $\bigcirc$  & $\bigcirc$ & $\bigcirc$  & $\bigcirc$ \\
\hline
Reversibility (${\rm DGT}_{\alpha}^{-1}$  exists) & $\bigcirc$  & $\bigcirc$  & $\bigcirc$   & $\bigcirc$ \\
\hline
Reversibility (${\rm DGT}_{\alpha}^{-1}={\rm DGT}_{-\alpha}$) & $\times$  & $\bigcirc$  & $\bigcirc$   & $\bigcirc$ \\
\hline
Additivity (${\rm DGT}_{\alpha_2}{\rm DGT}_{\alpha_1}={\rm DGT}_{\alpha_1+\alpha_2}$)& $\times$ & $\times$ & Approximate    & $\bigcirc$ \\
\hline
Singularities\textsuperscript{*} 
& $\alpha=k\pi$  & $\alpha=k\pi$ &  $\alpha=(2k+1)\pi$  & None \\
\hline
\multirow{2}{*}{Sampling intervals}&  \multirow{2}{*}{Arbitrary} & ${\Delta _x}{\Delta _v} = \frac{2\pi |\sin \alpha |}{N_1}$
&  $\Delta_u=\Delta_x$ & \hspace*{-.7cm}$ {\Delta _x} = {\Delta _y} $\\
  &  & ${\Delta _y}{\Delta _u} = \frac{2\pi |\sin \alpha |}{N_2}$
 & $ \Delta _v= \Delta _y$  & $  ={\Delta _u}={\Delta _v}= \sqrt {\frac{2\pi}{N}} $\\
\hline
Dominant complexity\textsuperscript{$*$} & Three 2D FFTs  & One 2D FFT  & Two 2D FFTs & Four matrix multiplications\textsuperscript{$\dag$}\\
\hline
\multicolumn{5}{l}{\textsuperscript{$*$}\footnotesize{For DGT-LCC and DGT-DFT with $\alpha\to k\pi$, the method in (\ref{eq:DGT208}) can avoid singularity problem, but the cost is one more 2D FFT. For DGT-CCC, }}\vspace{-3pt}\\
\multicolumn{5}{l}{\, \footnotesize{with $\alpha\to (2k+1)\pi$, the method in (\ref{eq:DGT216}) is used without complexity increase.}}\\
\multicolumn{5}{l}{\textsuperscript{$\dag$}\footnotesize{The two matrix multiplications in (\ref{eq:DGTdhgf48_1}) and two matrix multiplications in (\ref{eq:DGTdhgf48_3}) dominate the complexity of DGT-DHGF.}}
\end{tabular}
\end{center}
\end{table*}

\subsection{Accuracy}\label{subsec:Accuracy}
Next, we examine the accuracy of using the proposed DGTs to calculate the samples of continuous gyrator transform.
Consider a continuous input $g(x,y)$ and its gyrator transform is given by $G(u,v)$.
The accuracy of the DGTs is measured by the NRMSE (defined in (\ref{eq:ComProp18})) between ${G(p{\Delta _u},q{\Delta _v})}$ and $\textmd{DGT}_\alpha \left\{ {g(m{\Delta _x},n{\Delta _y})}\right\}$.
In Fig.~\ref{fig:Accuracy}, two examples are given.
In the first one, the input is a scaled Gaussian function $g(x,y)={e^{ - \frac{1}{2}s\left( {{x^2} + {y^2}} \right)}}$ with scaling parameter $s=0.4$.
Its closed-form gyrator transform is given by
\begin{align}\label{eq:ComAcc24}
\!\!G(u,v)\!=\!\frac{{e^{j\frac{1}{2} \cdot \frac{{({s^2} - 1)\sin 2\alpha }}{{{{\cos }^2}\alpha  + {s^2}{{\sin }^2}\alpha }}uv}}}{{\sqrt {{{\cos }^2}\alpha  + {s^2}{{\sin }^2}\alpha } }}\ {e^{ - \frac{1}{2} \cdot \frac{s}{{{{\cos }^2}\alpha  + {s^2}{{\sin }^2}\alpha }}\left( {{u^2} + {v^2}} \right)}}
\end{align}
according to \cite{rodrigo2007gyrator}.
For $N=101$, the $N\times N$ sampled scaled Gaussian, i.e. $g(m{\Delta _x},n{\Delta _y})$, with $\Delta_x=\Delta_y=\sqrt{2\pi/N}$ is depicted in Fig.~\ref{fig:Accuracy}(a).
The NRMSEs of the four proposed DGTs are calculated 
and illustrated in Fig.~\ref{fig:Accuracy}(b).
Notice that for the DGT-LCC, DGT-DFT and DGT-CCC, the method in (\ref{eq:DGT208}) or (\ref{eq:DGT216}) is used when $\alpha$ is close to the singularities.
We can find out that the DGT-CCC has the highest accuracy while the DGT-LCC has the lowest.
But generally speaking, all the DGTs have satisfactory performance in this example.
This is because the input signal has energy well concentrated around the origin of space/spatial-frequency planes, as shown in Fig.~\ref{fig:Accuracy}(a).
We consider an opposite example.
In (\ref{eq:DGTeig08}), it is mentioned that the RHGF of order $(k,l)$ is the eigenfunction of the gyrator transform with eigenvalue ${e^{ - j\alpha (k - l)}}$.
Fig.~\ref{fig:Accuracy}(c) shows the sampled RHGF of order $(25,40)$ with $\Delta_x=\Delta_y=\sqrt{2\pi/N}$ and $N=128$.
The accuracy of the DGT-LCC, DGT-DFT and DGT-CCC varies sharply  as the value of $\alpha$ changes.
This is because the energy of the input signal is not concentrate enough, as shown in Fig.~\ref{fig:Accuracy}(c).
Some steps of computation in these DGTs will result in aliasing (overlapping) effect.
On the contrary, the DGT-DHGF is much less affected by $\alpha$.
The accuracy of the DGT-DHGF mainly depends on the accuracy of DHGFs.
Since higher-order DHGFs are less accurate (see Fig.~\ref{fig:HGF_RMSE}), an input signal with more energy distributed on high-order DHGFs will yield lower accuracy.

\section{Applications}\label{sec:App}
A brief summary and comparisons of the four proposed DGTs are given in TABLE~\ref{tab:table1}.
For signal processing applications, the choice of the DGT depends on the sampling intervals of the input 2D signal.
If there are multiple options, generally speaking, the first choice is the DGT-CCC or the DGT-DHGF because of the additivity property.
Compared with the DGT-CCC, the DGT-DHGF has a little higher complexity but has perfect additivity property.
The second choice is the DGT-DFT 
because it has lower complexity then the DGT-LCC and the output size remains the same as the input.
If the sampling intervals do not satisfy any of the constraints of the DGT-DFT, DGT-CCC and DGT-DHGF, the DGT-LCC is recommended.
For most image processing applications, the sampling intervals are usually determined by oneself, and thus the DGT-CCC and DGT-DHGF are preferred.
In the following, we give a brief introduction of some applications of the DGTs.

\begin{figure}[t]
\centering
\includegraphics[width=\columnwidth,clip=true]{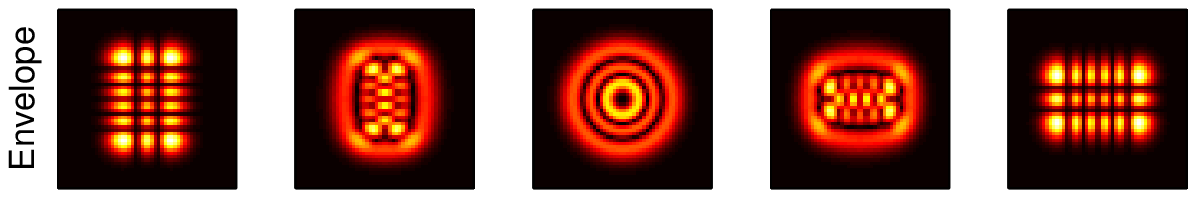}\\
\vspace{0.1cm}
\includegraphics[width=\columnwidth,clip=true]{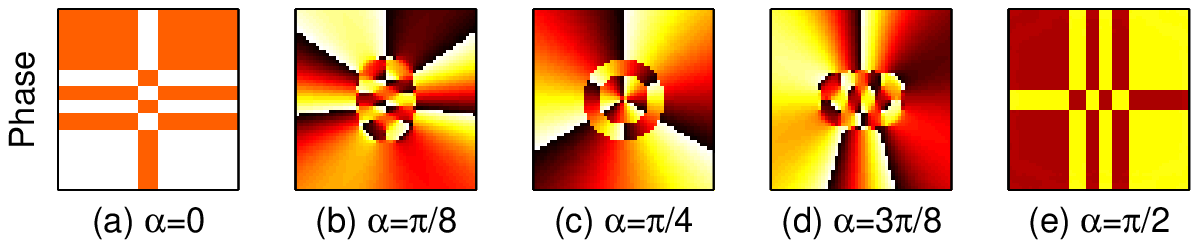}\\
\vspace{0.3cm}
\hspace{2pt}\includegraphics[width=0.9\columnwidth,clip=true]{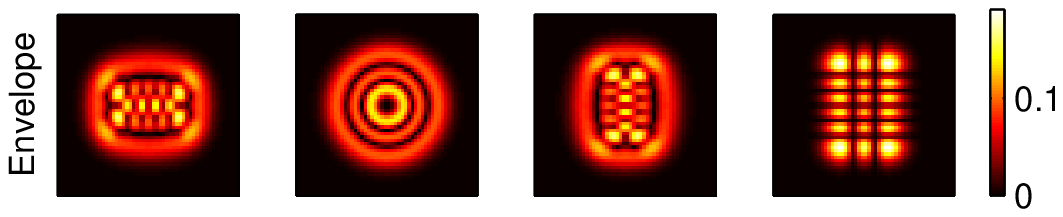}\\
\includegraphics[width=0.89\columnwidth,clip=true]{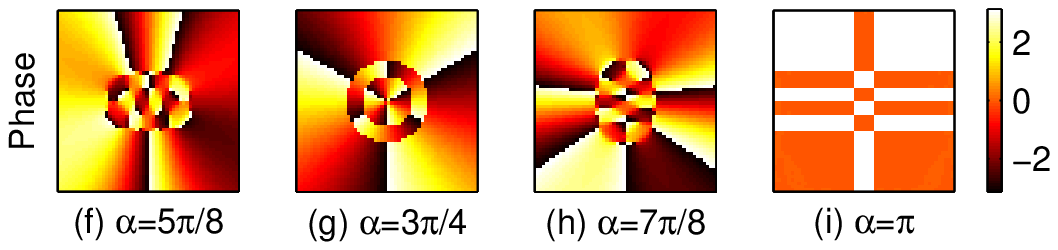}
\caption{
Magnitudes and phases of the DGT-LCCs of the $128\times128$ sampled HG mode $HG_{2,5}$ with ${\Delta _x}={\Delta _y}={\Delta _u}={\Delta _v}=\sqrt{2\pi/128}$:
(a) $\alpha=0$, (b) $\pi/8$, (c) $\pi/4$ (d) $3\pi/8$, (e) $\pi/2$, (f) $5\pi/8$, (g) $3\pi/4$, (h) $7\pi/8$ and (i) $\pi$.
The method in (\ref{eq:DGT208}) is used for $\alpha=0,\pi/8,7\pi/8,\pi$.
(Only display the outputs within $-17.9\leq u,v\leq17.9$.)
}
\label{fig:App_modeHG}
\end{figure}

\subsection{Mode Conversion}
One well-known application of the gyrator transform in optics is mode conversion \cite{rodrigo2007gyrator,rodrigo2007experimental}.
The gyrator transform can convert the Hermite Gaussian (HG) modes (i.e. 2D HGFs defined in (\ref{eq:DGTeig02}) and (\ref{eq:DGTeig04})) into the Laguerre Gaussian (LG) modes or other stable modes.
Since the HG modes are orthonormal to each other, the gyrator
transforms of the HG modes also form an orthonormal set.
Thus, these stable modes can be used for signal expansion and reconstruction.
Consider the input is $128\times128$ sampled HGF of order $(2,5)$, i.e. $HG_{2,5}(m\Delta_x,n\Delta_y)$ with $\Delta_x=\Delta_y=\sqrt{2\pi/128}$.
If we want to generate other stable modes with the same sampling intervals, i.e. ${\Delta _u}={\Delta _v}=\sqrt{2\pi/128}$, the DGT-LCC, DGT-CCC and DGT-DHGF are recommended.
In this simulation, the DGT-LCCs with $\alpha=0,\pi/8, 2\pi/8,\ldots,\pi$ are shown in Fig.~\ref{fig:App_modeHG}(a) to (i), respectively.
Note that when $\alpha$ is close to $k\pi$, i.e. $0$, $\pi/8$, $7\pi/8$ and $\pi$, 
the method in (\ref{eq:DGT208}) is used.
It is shown that the discrete LG modes can be obtained by the DGTs with $\alpha=\pi/4$ 
and $\alpha=3\pi/4$.

\begin{figure}[t]
\centering
\includegraphics[width=\columnwidth,clip=true]{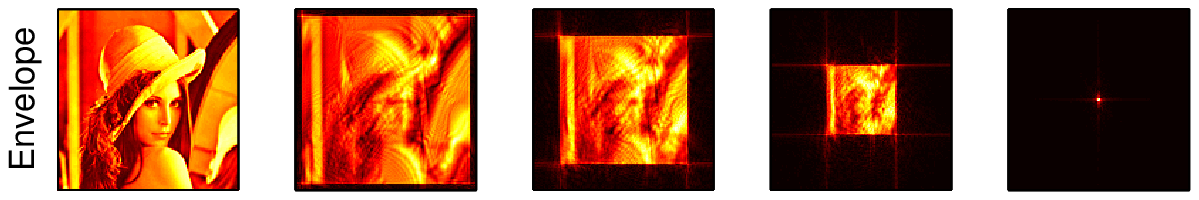}\\
\vspace{0.1cm}
\includegraphics[width=\columnwidth,clip=true]{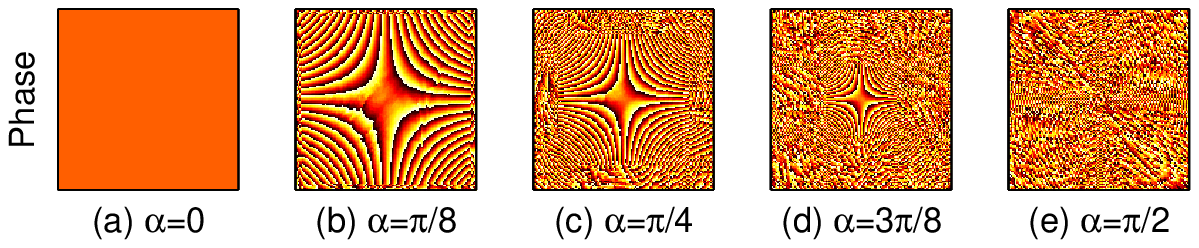}\\
\vspace{0.3cm}
\hspace{-20pt}\includegraphics[width=0.81\columnwidth,clip=true]{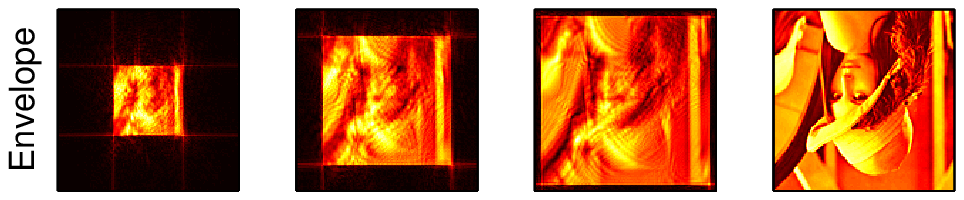}\\
\vspace{0.1cm}
\includegraphics[width=0.89\columnwidth,clip=true]{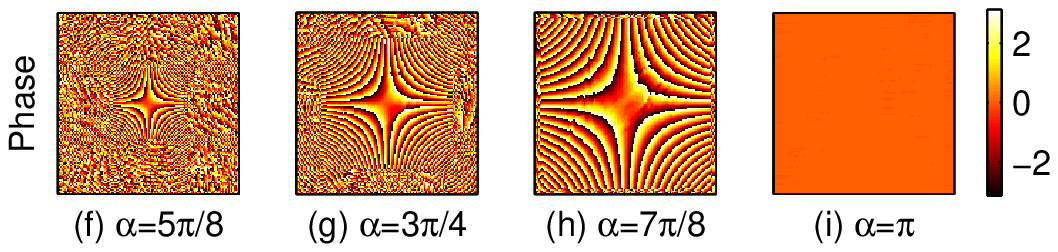}
\caption{
Magnitudes and phases of the DGT-CCCs of the $128\times128$ Lena image with ${\Delta _x}={\Delta _y}={\Delta _u}={\Delta _v}=\sqrt{2\pi/128}$:
(a) $\alpha=0$, (b) $\pi/8$, (c) $\pi/4$ (d) $3\pi/8$, (e) $\pi/2$, (f) $5\pi/8$, (g) $3\pi/4$, (h) $7\pi/8$ and (i) $\pi$.
The method in (\ref{eq:DGT216}) is used for $\alpha=5\pi/8,6\pi/8,7\pi/8,\pi$.
}
\label{fig:App_modeLena}
\end{figure}

Additionally, the DGT-CCCs of the $128\times128$ Lena image with $\alpha=0,\pi/8, 2\pi/8,\ldots,\pi$ are depicted in Fig.~\ref{fig:App_modeLena} as a reference.
Since the DGT-CCC is singular at $\alpha=\pi$, the method in (\ref{eq:DGT216}) is used when $\alpha=5\pi/8,6\pi/8,7\pi/8,\pi$.

\subsection{Sampling and Reconstruction}
\begin{figure}[t]
\centering
\includegraphics[width=\columnwidth, clip=true]{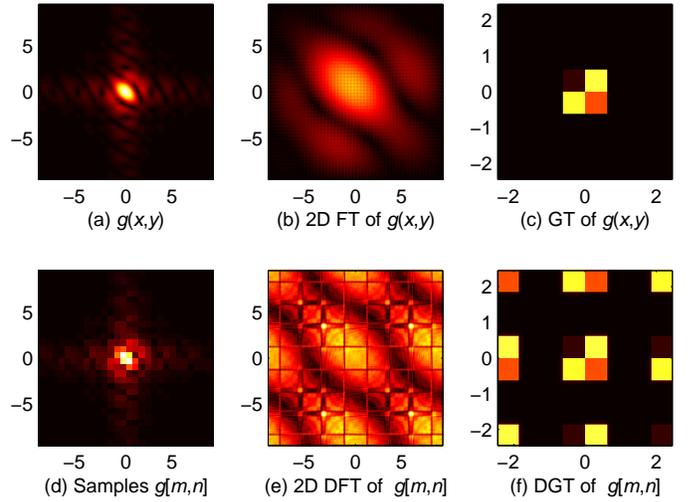}
\caption{
Magnitudes of (a) a continuous 2D signal $g(x,y)$, (b) 2D Fourier transform (FT) of $g(x,y)$, (c) gyrator transform (GT) of $g(x,y)$ with $\alpha=15^o$, (d) the samples $g[m,n]=g(0.666m,0.666n)$, (f) 2D DFT of $g[m,n]$, and (g) DGT-DFT of $g[m,n]$ with $\alpha=15^o$ and $\Delta_u=\Delta_v=0.0244$.
}
\label{fig:App_sampling}
\end{figure}

In \cite{pei2009properties}, the 2D signal sampling and reconstruction using the gyrator transform
are discussed.
It is possible that the given signal has smaller bandwidth in gyrator domain then in 2D Fourier domain, and  reconstruction in gyrator domain allows lower sampling rate.
Consider a 2D signal $g(x,y)$, the magnitude of which is shown in Fig.~\ref{fig:App_sampling}(a).
The 2D Fourier transform and the gyrator transform with angle $\alpha=15^o$ are shown in Fig.~\ref{fig:App_sampling}(b) and (c), respectively.
It can be found that $g(x,y)$ has much smaller bandwidth in gyrator domain, and thus lower sampling rate can be used if the reconstruction is performed in gyrator domain.
For example, consider that $g(x,y)$ is sampled with $\Delta_x=\Delta_y=0.666$ as shown in Fig.~\ref{fig:App_sampling}(d).
The 2D DFT depicted in Fig.~\ref{fig:App_sampling}(e) suffers from serious aliasing effect.
On the contrary, the DGT-DFT with $\alpha=15^o$ and $\Delta_u=\Delta_v=0.0244$ in Fig.~\ref{fig:App_sampling}(f) shows that perfect reconstruction can be done by placing a 2D lowpass mask in gyrator domain.
In practice, the optimal angle $\alpha$ may be unknown.
In this situation, the DGT-DHGF is superior due to its perfect additivity property.
One can iteratively perform the DGT-DHGF with some small angle 
until the output has the smallest aliasing effect.

\subsection{Watermarking}\label{subsec:water}
Roughly speaking, watermarking techniques can be classified into two categories, space domain and spatial-frequency domain.
The DFT,  discrete  cosine  transform (DCT) and  discrete  wavelet  transform (DWT) are some of the popular transforms used in spatial-frequency domain watermarking.
From Fig.~\ref{fig:App_modeLena}, the gyrator domain can be deemed as a joint space/spatial-frequency domain where the angle $\alpha$ determines the proportion of each domain.
Some watermarking schemes based on the gyrator transform have been proposed in \cite{singh2010digital,liu2010image,bhatnagar2011new,li2014optimized,yadav2015phase}.
The FRFT, introduced more than two decades before the gyrator transform, has been widely used in joint domain watermarking such as 
\cite{djurovic2001digital,feng2005multiple,yu2006digital,
nishchal2009optical,al2010digital,savelonas2010noise,
guo2011image,rawat2012blind,taba2013fractional}.
Since the 2D FRFT is highly related to the gyrator transform \cite{pei2009properties}, many works
of the 2D FRFT can be applied to the gyrator transform with similar performance.

\begin{figure}[t]
\centering
\includegraphics[width=\columnwidth, clip=true]{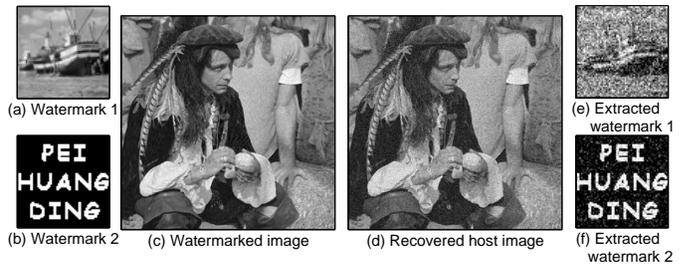}
\caption{
Watermarking based on DGT-CCC: (a) $64\times64$ watermark $W^{(1)}$, (b) $64\times64$ watermark $W^{(2)}$, (c) $256\times256$ watermarked image with PSNR 37.2dB, (d) recovered host image (PSNR 28.1dB) by removing the watermarks from the noisy watermarked image with white Gaussian noise (variance $\sigma^2=100$), (e) extracted watermark $W^{(1)}$ (PSNR 15dB) from the noisy watermarked image, and (f) extracted watermark $W^{(2)}$ (PSNR 17.6dB)  from the noisy watermarked image.
}
\label{fig:App_water1}
\end{figure}

For example, consider
the watermarking scheme based on \cite{djurovic2001digital,taba2013fractional}.
Given a host image $s[m,n]$, we calculate the 2D discrete FRFT (DFRFT) with angles $(\alpha,\alpha)$ and reorder the output coefficients into a nondecreasing sequence $S=\{S_l\left|\,|S_l|\geq|S_{l-1}|\right.\}$.
Next, two watermarks $W^{(1)}$ and $W^{(2)}$ are embedded in the coefficients with middle energy in order to avoid deformation on the watermarked image and attacks from low-pass filtering.
That is,
\begin{align}\label{eq:water04}
S_{l+Q}^{(w)} = \left\{ {\begin{array}{*{20}{l}}
{S_{l+Q}^{} + {k_1}W_{l}^{(1)} + j{k_2}W_{l}^{(2)},}&\ \ { 1\leq l\leq L}\\
{S_{l+Q}^{},}&\ \ \textmd{otherwise}
\end{array}} \right..
\end{align}
At last, the watermarked image is obtained by performing 2D DFRFT with angles $(-\alpha,-\alpha)$ on $S^{(w)}$.
The parameters $k_1$ and $k_2$ in (\ref{eq:water04}) are chosen to maintain high quality on the watermarked image.
This watermarking scheme can be applied to the gyrator transform by simply replacing the 2D DFRFT by DGT.
Fig.~\ref{fig:App_water1}(a) and (b) show two $64\times64$ watermarks, used as $W^{(1)}$ and $W^{(2)}$, respectively.
With $L=64^2$, $Q=8000$ and $k_1=k_2=0.15$, the $256\times256$ watermarked image obtained from DGT watermarking is depicted in Fig.~\ref{fig:App_water1}(c), where the PSNR is 37.2dB.
If the watermarked image suffers from white Gaussian noise with variance $\sigma^2=100$,
the recovered host image in Fig.~\ref{fig:App_water1}(d) and the extracted watermarks in  (e) and (f) have
PSNRs 28.1dB, 15dB and 17.6dB, respectively.
In this example, the DGT-CCC with $\Delta_x=\Delta_y=\Delta_u=\Delta_v=0.1567$ is utilized.
The results of the 2D DFRFT watermarking are similar to those of the DGT watermarking, having difference smaller then 0.3dB, and thus not shown here.

\begin{figure}[t]
\centering
\includegraphics[width=\columnwidth, clip=true]{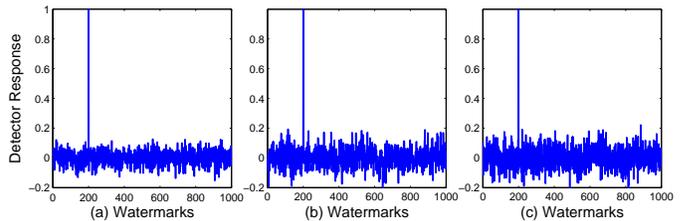}
\caption{
Normalized detector response over 1000 different sets of watermarks for a noisy watermarked image (noise variance $\sigma^2=100$): (a) detection in DGT domain, (b) detection in 2D DFRFT domain, and (c) detection after performing 1D DFRFT along vertical direction.
The 200th set is the correct set of watermarks.
The rest are generated by random integers within $[0,255]$.
}
\label{fig:App_water2}
\end{figure}

Next, we examine the performance of watermark detection for the noisy watermarked image.
The detection performance is measured by the detector response defined in \cite{djurovic2001digital}:
\begin{align}\label{eq:water08}
d = \sum\limits_{l = Q + 1}^{l = Q + L} {\left[ {W_{l - Q}^{(1)} - jW_{l - Q}^{(2)}} \right]} S_l^{(nw)},
\end{align}
where $S_l^{(nw)}$ denotes the 2D DFRFT/DGT coefficients of the noisy watermarked image.
The normalized detector
responses of the DGT and 2D DFRFT over 1000 different sets of watermarks are shown in Fig.~\ref{fig:App_water2}(a) and (b), respectively.
The 200th is the correct set of watermarks.
The rest are generated by random integers within $[0,255]$.
The detection in the DGT watermarking is somewhat more reliable than in the 2D DFRFT watermarking because the variance of detector response is smaller when incorrect watermarks are used.
Besides, since the 2D DFRFT is separable, it can also be implemented by two 1D DFRFTs along the vertical and horizontal directions, respectively.
It yields that one can try to detect the watermarks after performing only one 1D DFRFT.
Fig.~\ref{fig:App_water2}(c) shows the normalized detector response when detection is made after performing 1D DFRFT along vertical direction.
This implies
that the nonseparable transform, DGT, can provide higher security.


\subsection{Image Encryption}
One class of encryption techniques is to treat an image as
a data sequence and encrypt it by traditional ciphers such as DES, AES, IDEA and RC4. 
However, since images have some intrinsic features such as high redundancy and large size, 
other more efficient techniques such as chaotic mapping, pixel scrambling/shuffling and SCAN 
are used.
Plus, some of these techniques have been combined with the DFT, DCT and DWT for spatial-frequency domain encryption.

\begin{figure}[t]
\centering
\includegraphics[width=.95\columnwidth, clip=true]{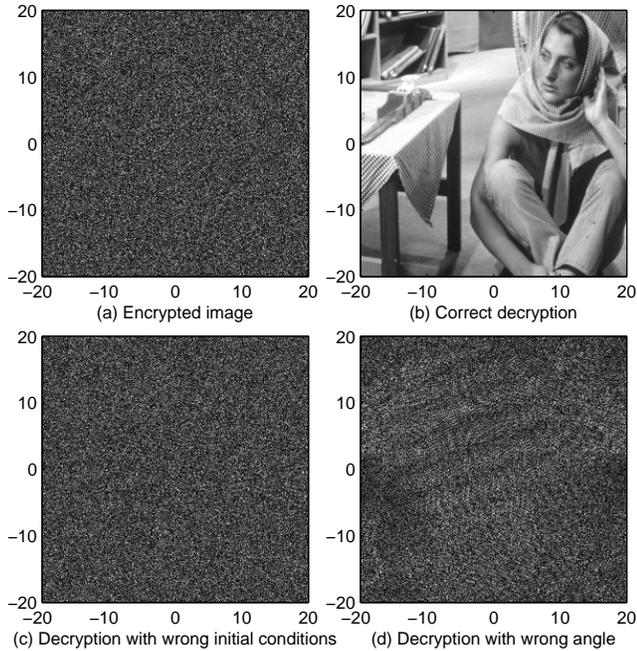}
\caption{
Image encryption based on DGT-DHGF with $\alpha=40^\circ$ and chaotic mapping with initial conditions within $[0,1]$:
(a) $256\times256$ encrypted image,
(b) decrypted image using correct initial conditions and angle,
(c) decrypted image using wrong initial conditions with very small errors $\pm10^{-12}$,
and (d) decrypted image using wrong angle with a very small error $0.0001^{\circ}$.
}
\label{fig:App_encrypt_1}
\end{figure}

Image encryption in joint space/spatial-frequency domain has also attracted increasing attentions in recent years.
In gyrator domain, a number of encryption schemes based on random phase encoding, chaotic mapping, phase retrieval algorithm, Arnold transform and/or pixel scrambling have been proposed
\cite{rodrigo2007applications,li2008double,li2009image,singh2009gyrator,
khanzadi2010image,liu2010image2,liu2010double,
abuturab2012securing,abuturab2012color,
abuturab2013color,liu2013double,li2013double}.
A review of encryption techniques in fractional Fourier domain and gyrator domain is available in \cite{liu2014review,khan2014literature}.
In the following, we give an example of gyrator domain encryption based on \cite{khanzadi2010image}.
The encryption scheme consists of four steps:
\begin{itemize}[\labelindent=4pt]
\item[1.] Calculate the DGT of the input image with angle $\alpha$.
\item[2.] Represent each coefficient of the DGT by $K$ bits.
\item[3.] Encrypt the $k$-th bits of all the coefficients by chaotic maps with initial conditions within $[0,1]$, and repeat the process for $k=1,2,\ldots,K$.
\item[4.] Obtain the encrypted image by performing inverse DGT (i.e. DGT with $-\alpha$) to the encrypted coefficients.
\end{itemize}
Note that the scheme in \cite{khanzadi2010image} lacks the 4th step.
Fig.~\ref{fig:App_encrypt_1}(a) shows the $256\times256$ encrypted image through DGT-DHGF with $\alpha=40^\circ$ and chaotic mapping with $K=16$.
The decrypted image using correct initial conditions and correct angle is depicted in Fig.~\ref{fig:App_encrypt_1}(b).
Fig.~\ref{fig:App_encrypt_1}(c) and (d) show the decrypted images using wrong initial conditions with very small errors $\pm10^{-12}$ and using wrong angle with a very small error $0.0001^{\circ}$, respectively.
This example shows why the angle of the gyrator transform (or 2D FRFT) is regarded as a secrete key in some papers.

\begin{figure}[t]
\centering
\includegraphics[width=\columnwidth, clip=true]{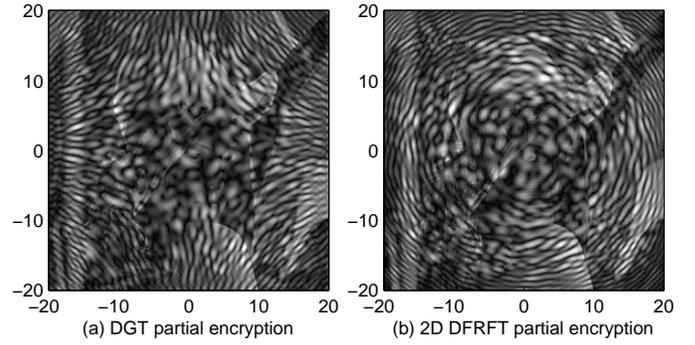}
\caption{
$256\times256$  partial encrypted images obtained by encrypting only the central $28\times28$ coefficients
(a) in the DGT domain with $\alpha=70^\circ$
and (b) in the 2D DFRFT domain with $\alpha=70^\circ$.
}
\label{fig:App_encrypt_2}
\end{figure}

Compared with ciphers and encryption techniques, the gyrator transform has minor contributions to resistant against attacks because it is linear.
Despite this, the gyrator transform has some benefits to image encryption such as:
\begin{itemize}[\labelindent=4pt]
\item The gyrator transform has energy compaction property (see the cases of $\alpha$ close to $\pi/2$ in Fig.~\ref{fig:App_modeLena}).
     So performing encryption only on the high-energy part can achieve lower complexity with good enough security.
\item Multiple encryption stages operating in different gyrator domains (different angles) may yield higher security than in the same domain. For example, performing random phase encoding multiple times in the same domain is equivalent to just once.
\item Partial encryption in gyrator domain enables information to be secured with different levels of security for different needs. An example is presented below.
\end{itemize}
With $256\times256$ Lena image as the input, use the encryption scheme mentioned in the previous paragraph again except that  in the 3rd step only the central $28\times28$ coefficients are encrypted.
The encrypted image using DGT-DHGF with $\alpha=70^\circ$ is shown in Fig.~\ref{fig:App_encrypt_2}(a).
This partial encryption is similar to the combination of low-frequency part encryption in spatial-frequency domain and central region encryption in space domain.
The value of $\alpha$ can be used to control the security levels of encryption in these two domains.
Therefore, the central region of Fig.~\ref{fig:App_encrypt_2}(a) suffers from space domain encryption and low-frequency encryption while the marginal zone only suffers from the low-frequency encryption.
Replacing the DGT by the 2D DFRFT, the result of partial encryption in fractional Fourier domain, depicted in Fig.~\ref{fig:App_encrypt_2}(b), provides somewhat different encryption effect for different needs.

\section{Conclusion}\label{sec:Con}
In this paper, we develop the 
DGTs based on the 
2D LCT and
based on the eigenfunctions of the gyrator transform.
The parameter matrix that makes the 2D LCT equivalent to the gyrator transform is presented.
Based on the decompositions of the parameter matrix, three kinds of DGTs are developed.
The constraints, properties and computational algorithms of these DGTs are discussed.
These DGTs have singularities at $\alpha=k\pi$ or $\alpha=(2k+1)\pi$.
Therefore, we propose a method that makes these DGTs avoid their singularities and still useful when $\alpha$ is close to $k\pi$ or $(2k+1)\pi$.
The 4th kind of DGT is based on the $45^\circ$ counterclockwise rotation of the 
2D HGFs, which are the eigenfunctions of the gyrator transform.
An efficient computational algorithm for this DGT is developed.
The advantage of this DGT is the perfect additivity property, which makes it superior in many applications.
We also give a brief introduction to some important applications of the proposed DGTs, including mode conversion, sampling and reconstruction, 
watermarking and image encryption.

\appendices
\section{Rotated Hermite Gaussian Functions (RHGFs) and Gyrator Transform}\label{App:RHGF}
In \cite{wunsche2000general}, it has been shown that
\begin{align}\label{eq:AppRHGF04}
&H{G_{k,l}}(x\cos \phi  - y\sin \phi ,x\sin \phi  + y\cos \phi )\nn\\
 &= \sum\limits_{s = 0}^L {\sqrt {\frac{{s!(L - s)!}}{{k!l!}}} {{( - \sin \phi )}^{k - s}}{{(\cos \phi )}^{l - s}}} \nn\\
&\qquad\qquad\qquad \cdot P_s^{(k - s,l - s)}\left[ {\cos (2\phi )} \right]H{G_{s,L - s}}(x,y),
\end{align}
where $L=k+l$ and $P_s^{(\beta,\gamma)}$ is the Jacobi polynomial.
From the definition of the Wigner d-function $d^J_{M_1M_2}(\beta)$ in \cite{varshalovich1988quantum}, 
the coefficient of $H{G_{s,L - s}}(x,y)$ in (\ref{eq:AppRHGF04}) is equal to $d_{\frac{{l - k}}{2},\frac{L}{2} - s}^{\frac{L}{2}}(  -2\phi )$.
The RHGF defined in (\ref{eq:DGTeig06}) is the $45^\circ$ counterclockwise rotation of the 2D HGF, i.e. $\phi=-\pi/4$.
Replacing the coefficients in (\ref{eq:AppRHGF04}) by $d_{\frac{{l - k}}{2},\frac{L}{2} - s}^{\frac{L}{2}}(  \pi/2 )$, relation (\ref{eq:DGTeig16}) is proved.

Substituting (\ref{eq:DGTeig16}) into  (\ref{eq:DGTeig12}) leads to
\begin{align}\label{eq:AppRHGF10}
{{\widehat g}_{k,l}} = \sum\limits_{s = 0}^L {d_{\frac{{l - k}}{2},\frac{L}{2} - s}^{\frac{L}{2}}\left( { \frac{\pi }{2}} \right)} {{\widetilde g}_{s,L-s}},
\end{align}
where ${{\widetilde g}_{s,L-s}}$ is defined in (\ref{eq:DGTeig20}).
Substituting (\ref{eq:DGTeig16}) and (\ref{eq:AppRHGF10}) into
(\ref{eq:DGTeig14}) yields that
\begin{align}\label{eq:AppRHGF12}
G(u,v) &= \sum\limits_{k = 0}^\infty  {\sum\limits_{l = 0}^\infty  {{e^{ - j\alpha (k - l)}}\left[ {\sum\limits_{r = 0}^L {d_{\frac{{l - k}}{2},\frac{L}{2} - r}^{\frac{L}{2}}\left( {\frac{\pi }{2}} \right)} {{\widetilde g}_{r,L - r}}} \right]} }  \nn\\
&\qquad\quad\cdot\sum\limits_{s = 0}^L {d_{\frac{{l - k}}{2},\frac{L}{2} - s}^{\frac{L}{2}}\left( {\frac{\pi }{2}} \right)} H{G_{s,L - s}}(u,v),
\end{align}
where $L=k+l$. Assume $k' = s$, $l' = L - s$, and then the above equation can be rewritten as
\begin{align}\label{eq:AppRHGF14}
G(u,v) &= \sum\limits_{k' = 0}^\infty  {\sum\limits_{l' = 0}^\infty  {\sum\limits_{r = 0}^L {\left[ {\sum\limits_{l = 0}^L {d_{l - \frac{L}{2},\frac{{l' - k'}}{2}}^{\frac{L}{2}}\left( {\frac{\pi }{2}} \right)} } d_{l - \frac{L}{2},\frac{L}{2} - r}^{\frac{L}{2}}\left( {\frac{\pi }{2}} \right)\right.} } }   \nn\\
&\qquad\qquad\cdot\left. {{e^{ - j\alpha (L - 2l)}}} \right]{{\widetilde g}_{r,L - r}}H{G_{k',l'}}(u,v),
\end{align}
where $L=k'+l'$.
In \cite{trapani2006calculation}, it has been mentioned that
\begin{align}\label{eq:AppRHGF16}
\!\!\!\sum\limits_{M =  - J}^{ + J}\!\!\!\! {d_{M,{M_1}}^J\!\!\left( \!{\frac{\pi }{2}} \right)\!\!d_{M,{M_2}}^J\!\!\left(\! {\frac{\pi }{2}} \right)\!\!{e^{jM\beta }}} \!= \!D_{{M_1},{M_2}}^J\!\!\left(\! { - \frac{\pi }{2},\beta ,\frac{\pi }{2}} \right)\!,
\end{align}
where the definition of $D_{{M_1},{M_2}}^J$ has been shown in (\ref{eq:DGTeig22}).
Therefore, in (\ref{eq:AppRHGF16}), let $\beta  = 2\alpha$, $J = \frac{L}{2}$, $M = l - \frac{L}{2}$, ${M_1} = \frac{{l' - k'}}{2}$, and ${M_2} = \frac{L}{2} - r$, and then
(\ref{eq:DGTeig18}) is proved.

\bibliographystyle{IEEEtran}

\begin{IEEEbiography}[{\includegraphics[width=1in,height=1.25in,clip,keepaspectratio]{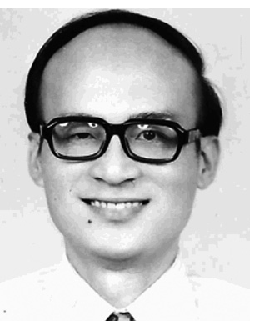}}]
{Soo-Chang Pei} (SM'89-F'00-LF'15) was born in Soo-Auo, Taiwan, in 1949. He received the B.S.E.E. degree from National Taiwan University, Taipei, Taiwan, in 1970, and the M.S.E.E. and Ph.D. degrees from the University of California Santa Barbara, Santa Barbara, in 1972 and 1975, respectively. From 1970 to 1971, he was an Engineering Officer with the Chinese Navy Shipyard. From 1971 to 1975, he was a Research Assistant with the University of California Santa Barbara. He was a Professor and the Chairman of the Department of Electrical Engineering with the Tatung Institute of Technology, Taipei, from 1981 to 1983 and with National Taiwan University from 1995 to 1998. From 2003 to 2009, he was the Dean of the College of Electrical Engineering and Computer Science with National Taiwan University. He is currently a Professor with the Department of Electrical Engineering, National Taiwan University. His research interests include digital signal processing, image processing, optical information processing, and laser holography.

Dr. Pei was a recipient of the National Sun Yet-Sen Academic Achievement Award in Engineering in 1984, the Distinguished Research Award from the National Science Council from 1990 to 1998, the Outstanding Electrical Engineering Professor Award from the Chinese Institute of Electrical Engineering in 1998, the Academic Achievement Award in Engineering from the Ministry of Education in 1998, the Pan Wen-Yuan Distinguished Research Award in 2002, and the National Chair Professor Award from the Ministry of Education in 2002. He was the President of the Chinese Image Processing and Pattern Recognition Society in Taiwan from 1996 to 1998 and is a member of Eta Kappa Nu and the Optical Society of America. He became an IEEE Fellow in 2000 for his contributions to the development of digital eigenfilter design, color image coding and signal compression and to electrical engineering education in Taiwan.
\end{IEEEbiography}

\begin{IEEEbiography}[{\includegraphics[width=1in,height=1.25in,clip,keepaspectratio]{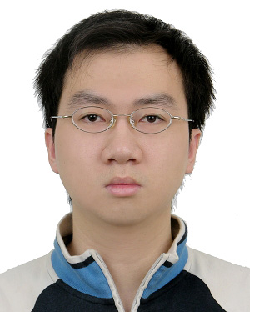}}]
{Shih-Gu Huang} was born in Taiwan in 1984. He received the B.S. degree in electrical engineering and the M.S. degree in communications engineering from National Tsing Hua University, Hsinchu, Taiwan, in 2007 and 2009, respectively. He is currently working toward the Ph.D. degree in the Graduate Institute of Communication Engineering, National Taiwan University, Taipei, Taiwan. His research interests include digital signal processing, time-frequency analysis, fractional Fourier transform, and linear canonical transform.
\end{IEEEbiography}

\begin{IEEEbiography}[{\includegraphics[width=1in,height=1.25in,clip,keepaspectratio]{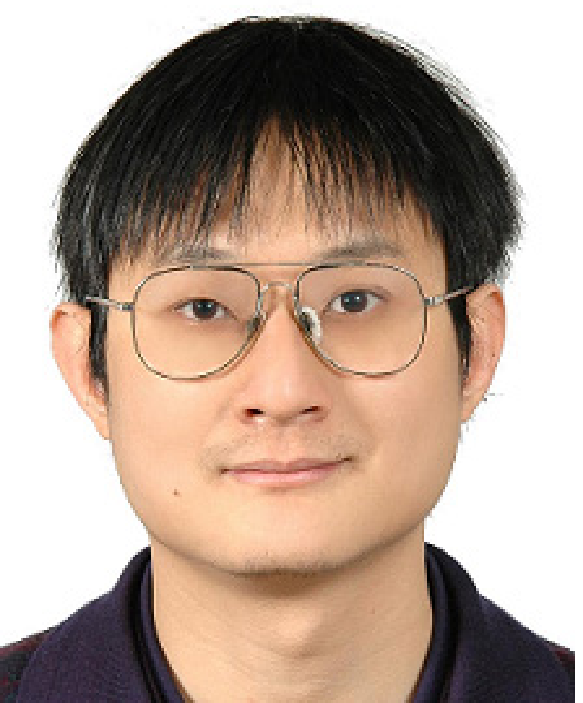}}]
{Jian-Jiun Ding} was born in 1973 in Taiwan. He received a B.S. degree in 1995, a M.S. degree in 1997, and a Ph.D. degree in 2001, all in electrical engineering from the National Taiwan University (NTU), Taipei, Taiwan. From 2001 to 2006, he was a postdoctoral researcher. From 2006 to 2012, he was an assistant professor with the Department of Electrical Engineering of NTU.
He is currently an associate professor with the Department of Electrical Engineering, NTU. He is also a senior member of IEEE. His current research areas include time-frequency analysis, fractional Fourier transforms, linear canonical transforms, wavelet transforms, image processing, image compression, orthogonal polynomials, fast algorithms, integer transforms, quaternion algebra, pattern recognition, filter design, etc.
\end{IEEEbiography}

\vfill

\end{document}